\documentclass[final,5p,times,twocolumn]{elsarticle}

\usepackage{amsmath,amsfonts,amssymb}
\usepackage{algorithmic}
\usepackage{algorithm}
\usepackage{array}
\usepackage{textcomp}
\usepackage{stfloats}
\usepackage{url}
\usepackage{verbatim}
\usepackage{multirow}
\usepackage{adjustbox}
\usepackage{graphicx}
\usepackage{float} 
\usepackage{hyperref}
\usepackage{booktabs}
\usepackage{pifont}
\usepackage{xcolor}
\usepackage{caption}
\usepackage{subcaption}
\usepackage{amssymb}
\usepackage{amsmath}

\usepackage{CJKutf8} 

\hypersetup{
    colorlinks=true,
    linkcolor=blue,
    filecolor=magenta,      
    urlcolor=cyan,
}

\journal{Computer-Aided Design} 

\begin{document}

\begin{frontmatter}

\title{Test-time Adaptation for Point Cloud Denoising under Unseen Noise}


\author[1]{Kun Jiang}
\ead{jiangkun532@qq.com}

\author[1]{Shida Fu}
\ead{940399011@qq.com}

\author[1]{Fanhao Yi}
\ead{yifanhao\_njust@qq.com}

\author[2]{Weiqing Li}
\ead{li\_weiqing@njust.edu.cn}

\author[1]{Zhiyong Su\corref{cor1}}
\ead{su@njust.edu.cn}

\cortext[cor1]{Corresponding author.}

\affiliation[1]{organization={School of Automation, Nanjing University of Science and Technology},
            city={Nanjing},
            state={Jiangsu Province},
            postcode={210094}, 
            country={P.R. China}}

\affiliation[2]{organization={School of Computer Science and Engineering, Nanjing University of Science and Technology},
            city={Nanjing},
            state={Jiangsu Province},
            postcode={210094}, 
            country={P.R. China}}

\begin{abstract}
Although deep learning-based point cloud denoising methods have demonstrated impressive results in recent years, their effectiveness usually degrades substantially in real-world scenarios where the noise distribution of test point clouds deviates from that of the training set.
To address this challenge, we propose a test-time adaptive denoising (TTAD) framework that enables existing denoising approaches to adapt to unknown noise patterns at test time without relying on additional supervision or retraining from source data.
Taking a pre-trained denoising model and a test noisy point cloud as input, TTAD first employs a residual-driven noise modeling (RDNM) module to estimate the sample-specific noise distribution through analyzing the deviation between the noisy input and its pseudo-clean version denoised by the initial model.
Then, a noise-aware contrastive adaptation (NACA) module is designed to build sample-specific noisy–noisier test-time training pair from the denoised point cloud, and update the model parameters over a few adaptation steps per sample at test time through contrastive learning.
Furthermore, TTAD naturally extends to a continual TTAD (CTTAD) setting, facilitating efficient progressive refinement across sequential test samples with consistent noise distributions, while reducing the computational cost per sample.
Extensive experiments on both synthetic and real-world datasets demonstrate that our approach significantly improves the generalization ability of existing denoising models under unseen noise, achieving superior performance across multiple evaluation metrics.
The source code is released at \href{https://github.com/zuowenzhitiao/TTAD}{https://github.com/zuowenzhitiao/TTAD}.
\end{abstract}


\end{frontmatter}


\section{Introduction}
\label{sec:intro}

\begin{figure}[!t]
    \centering
    \includegraphics[width=0.48\textwidth]{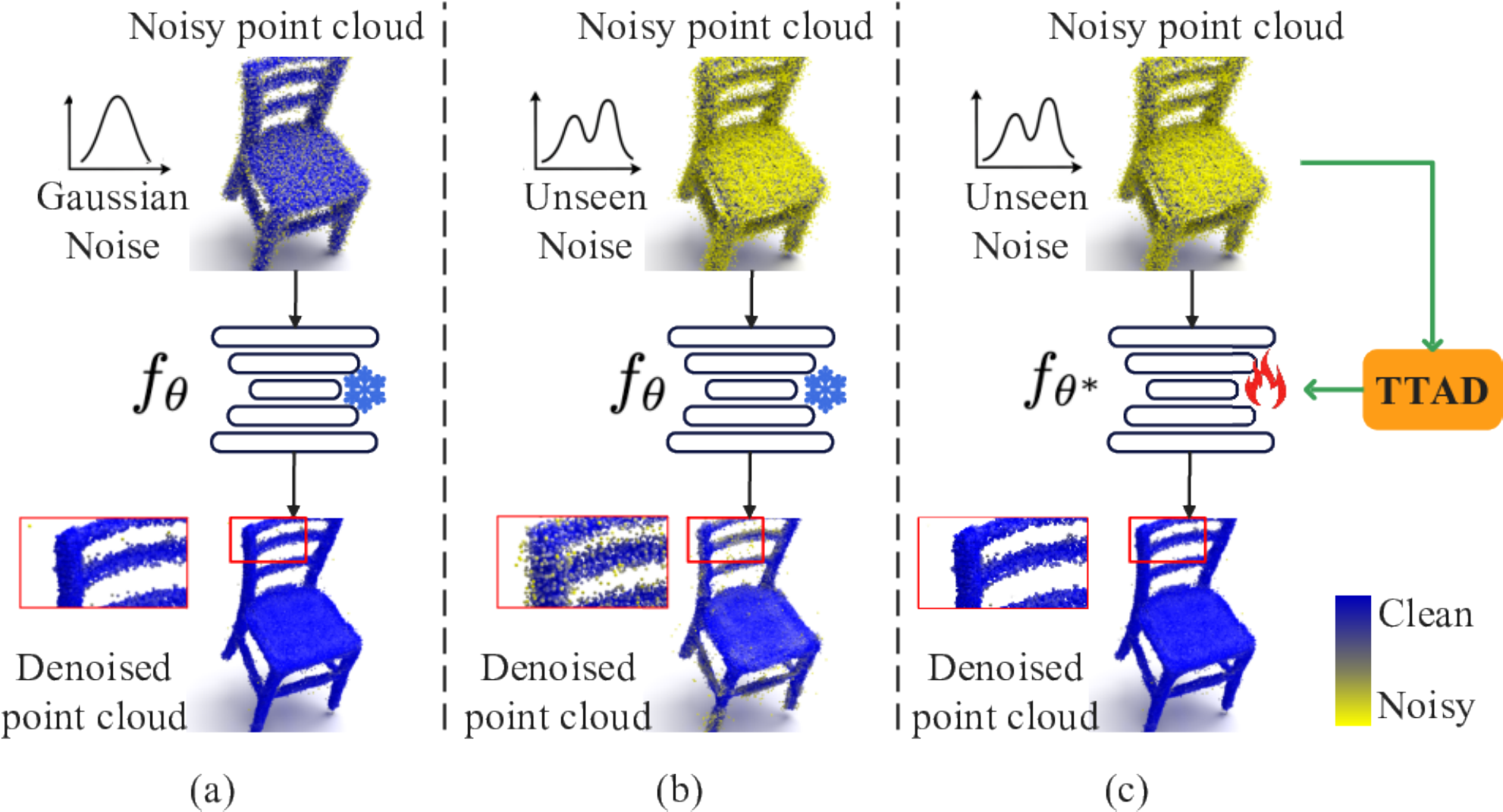} 
    \caption{Illustration of existing point cloud denoising methods and the proposed test-time adaptive denoising (TTAD) framework.
            The source model (PointFilter~\cite{zhang2020pointfilter}) $f_{\theta}$ is trained on synthetic datasets with Gaussian noise. $f_{\theta}$ performs well on noisy point clouds with seen (Gaussian) noise (a), but fails under unseen noise (b). The proposed TTAD framework significantly enhances the denoising performance of $f_{\theta}$ by adapting it to unseen noise distributions during test time (c).}
    \label{fig:problem_show}
\end{figure}

Point cloud is a fundamental representation of 3D geometric data, with widespread applications in autonomous driving \cite{luo2021self,10816198}, industrial inspection \cite{nguyen2018comparison}, digital twins \cite{xue2020lidar}, etc.
However, in practice, they are often corrupted by noise due to sensor limitations, environmental interference, or pose estimation errors, significantly impairing downstream tasks such as surface reconstruction, registration, and object recognition \cite{9723853,cao2022raw,fugacci2024reconstruction}.
Recent advances in deep learning-based point cloud denoising have demonstrated promising results \cite{zhou2022point,9903481,10099466}.
These methods are typically trained on synthetic datasets, either in a supervised \cite{9903481,10099466,luo2021score,luo2020differentiable,zhang2020pointfilter,de2023iterativepfn,rakotosaona2020pointcleannet} or an unsupervised fashion \cite{hermosilla2019total,su2025sitf,wang2024noise4denoise}.
They achieve state-of-the-art performance when the test noisy point clouds (target) and the training point clouds (source) belong to the same distribution.
However, the performance of these methods is often substantially degraded at test time when there exists a noise distribution gap between the training and test point clouds, which is termed "domain shift" problem, as illustrated in Figure~\ref{fig:problem_show}.

Test-time adaptation (TTA) \cite{liang2025comprehensive} has emerged as a promising solution to improve model generalization against domain shift through acclimating a trained model from the source domain to unlabeled test data during the test phase.
Recently, some pioneering studies have explored the TTA technique for image denoising \cite{mohan2021adaptive,vaksman2020lidia, fahim2023ss}.
These methods typically introduce auxiliary self-supervised objectives at test time, enabling adaptation using only unlabeled samples.
For example, Lidia~\cite{vaksman2020lidia} performs test-time adaptation by injecting synthetic noise into the denoised output and optimizing a denoising loss.
SS-TTA~\cite{fahim2023ss} employs Gaussian noise to create sample pairs for contrastive learning.
However, they all rely on sample-agnostic and synthetic noises (e.g., Gaussian noise), which may not accurately reflect the corruption patterns present in real-world test data.
This mismatch often leads to suboptimal adaptation, especially in denoising scenarios where noise characteristics can be highly diverse, sample-specific, and structurally complex.
Therefore, exploring the application of TTA to point cloud denoising represents a critical yet challenging research direction for improving generalization ability of current point cloud denoising methods, particularly in real-world noisy point cloud denoising scenarios.

In this paper, we propose a test-time adaptation framework termed test-time adaptive denoising (TTAD), which empowers a pre-trained point cloud denoising model to dynamically adapt to unseen noise during the test phase, without access to clean supervision or source-domain data.
Given a pre-trained point cloud denoising model, TTAD first introduces a residual-driven noise modeling (RDNM) module to capture sample-specific noise pattern in an unsupervised manner by estimating the error field between the noisy input and its pseudo-clean prediction.
Then, a noise-aware contrastive adaptation (NACA) module is designed to update the parameters of the pre-trained denoising model over a few adaptation steps per sample at test time in a self-supervised manner.
NACA achieves adaptation by constructing sample-specific noisy–noisier input pair and enforcing consistency via contrastive learning.
To further stabilize the adaptation process, exponential moving average (EMA) is incorporated to prevent overfitting to individual noise pattern.
Finally, our TTAD framework allows the pre-trained denoising model to achieve sample-wise test-time adaptation without network modifications.
Furthermore, for scenarios where point clouds encounter noise distributions with consistent characteristics, we extend our TTAD to a continual TTAD (CTTAD) setting, where the denoising model incrementally evolves across test samples by reusing previously adapted parameters.
This formulation not only reduces computational overhead by avoiding repeated adaptation for each input, but also improves stability through gradual knowledge accumulation across test samples.
The key contributions of this paper are summarized as follows:
\begin{itemize}
    \item We present a systematic study of test-time adaptation for point cloud denoising under unseen noise.
    To this end, we propose a general framework, termed TTAD, which allows pre-trained denoising models to achieve sample-wise test-time adaptation under unseen noise during the test phase.
    The TTAD framework can be seamlessly integrated with a wide range of existing point cloud denoising methods, without altering their original network architectures.

    \item We propose a noise-aware contrastive adaptation module that performs self-supervised parameter updates on the pre-trained denoising model during test time via contrastive learning.
    Specifically, it leverages a residual-driven noise modeling mechanism to construct sample-specific noisy–noisier training pair, enabling the model to iteratively refine its parameters in response to the noise characteristics of the current sample.

    \item We further extend TTAD to continual TTAD, facilitating efficient and scalable adaptation across sequential test noisy point clouds with consistent noise characteristics, while reducing per-sample computational cost, which makes the method suitable for real-world deployment in streaming applications.

    \item Extensive experiments conducted on both synthetic and real-world datasets demonstrate that TTAD and CTTAD consistently improve the performance of state-of-the-art denoising models under diverse unseen noise conditions, achieving notable gains across multiple evaluation metrics.
\end{itemize}

\section{Related Works} 
\label{sec:related_works}

\subsection{Deep Learning-based Point Cloud Denoising}

Deep learning-based point cloud denoising methods can be broadly categorized into supervised and unsupervised approaches, depending on whether clean ground-truth data is used during training.

\subsubsection{Supervised Point Cloud Denoising}

Supervised point cloud denoising methods are typically trained on synthetic clean-noisy point cloud pairs, where degradation is commonly simulated by adding Gaussian noise to clean point cloud.
These approaches can be broadly classified into three categories: 1) resampling, 2) probabilistic modeling, and 3) displacement prediction.
In resampling-based methods, denoising is achieved through downsampling and upsampling reconstruction on low-noise surfaces, yet geometric details may be lost in high-curvature regions.
For example, DMRDenoise \cite{luo2020differentiable} introduces differentiable pooling to downsample points to a low-noise manifold and then upsamples them back to a dense output.
To address the computational cost of dense grids, modern methods leverage sparse convolutional backbones via frameworks like Minkowski Engine \cite{8953494} and TorchSparse \cite{tang2022torchsparse}. These approaches typically utilize 3D U-Net architectures \cite{CABR16} to extract multi-scale representations from sparse voxels for effective resampling. For instance, Choe et al. \cite{choe2022deep} employed a sparse stacked hourglass network for voxel-space denoising and densification.
Probabilistic modeling methods, such as ScoreDenoise \cite{luo2021score} and its extension DeepPSR \cite{chen2022deep}, formulate denoising as an iterative inference process using score-based generative modeling, often requiring computationally intensive Langevin sampling.
Displacement prediction approaches (e.g., PointCleanNet~\cite{rakotosaona2020pointcleannet}, which is based on PointNet, and PointFilter~\cite{zhang2020pointfilter}) learn point-wise offsets toward clean surfaces.
Recent advances, such as IterativePFN~\cite{de2023iterativepfn}, adopt iterative refinement modules to jointly optimize displacement prediction and confidence estimation.
Moreover, recent methods explore denoising by disentangling noise in the latent space.
PDFlow \cite{mao2022pd} leverages normalizing flows to decouple noise from the underlying structure in a high-dimensional latent space.
Similarly, PD-LTS \cite{Mao_2024_CVPR} employs graph convolution and invertible neural networks for latent-space denoising.

However, supervised denoising methods often suffer from significant performance degradation when the noise distribution at test time deviates from the synthetic noise used during training.
This distribution gap limits their generalization to unseen noise types, particularly in test-time scenarios where the noise characteristics are unknown and vary considerably across domains.

\subsubsection{Unsupervised Point Cloud Denoising}

Unsupervised point cloud denoising methods aim to train denoising models without access to clean ground-truth data, typically by exploiting the internal structure of point clouds or constructing surrogate supervision signals.
TotalDenoising \cite{hermosilla2019total} leverages intrinsic structural information of point clouds to provide self-supervision for model training.
In contrast, methods such as SITF \cite{su2025sitf} and Noise4Denoise \cite{wang2024noise4denoise} construct noisy-noisier training pairs by injecting Gaussian noise into input point clouds.

While these approaches benefit from the unsupervised paradigm and generally exhibit good cross-domain generalization, their performance is often limited by the absence of clean supervision, resulting in suboptimal denoising quality.
Moreover, methods that construct noisy-noisier pairs (e.g., SITF~\cite{su2025sitf} and Noise4Denoise~\cite{wang2024noise4denoise}) typically rely on synthetic Gaussian noise injection, which does not account for the actual distribution of test-time noise.

\subsection{Test-Time Adaptation and Contrastive Learning}

\subsubsection{Test-Time Adaptation for Point Clouds} 
Test-time adaptation (TTA) \cite{liang2025comprehensive} has emerged as a promising solution to improve model generalization against domain shift by adapting a trained model to unlabeled test data during inference.
While TTA has been extensively studied in image restoration \cite{mohan2021adaptive,vaksman2020lidia,fahim2023ss}, its application to 3D point clouds remains relatively underexplored. 
Existing point-cloud TTA studies mainly focus on high-level tasks such as classification, segmentation, and open-vocabulary analysis. 
For example, Wang et al.~\cite{wang2024bftta} propose a backpropagation-free two-stream framework for efficient 3D test-time adaptation, and Zou et al.~\cite{zou2024hgl} introduce hierarchical geometry learning to exploit local, global, and temporal geometric cues for point cloud segmentation. 
More recently, diffusion-based methods have also been explored for point-cloud TTA. CloudFixer~\cite{shim2024cloudfixer} performs test-time input adaptation via diffusion-guided geometric transformations, while Dastmalchi et al.~\cite{dastmalchi2025test} propose a denoising-diffusion-based adaptation framework in latent space. 
LaPDA~\cite{LaPDA2025} improves point cloud denoising through adaptive latent-space modeling, providing a more structured representation for separating geometry from noise.

\subsubsection{Contrastive Learning for Point Clouds} 
Contrastive learning has shown strong potential for self-supervised 3D representation learning.
PointContrast~\cite{Xie_2020_ECCV} highlights the effectiveness of enforcing feature consistency across transformed views for point cloud understanding. 
Later works further extend contrastive learning to cross-modal alignment~\cite{afham2022crosspoint}, efficient point cloud representation learning~\cite{li2023tothepoint}, and large-scale scene-level pre-training~\cite{wu2023msc}. 
Inspired by these advances, we introduce a noise-aware contrastive adaptation strategy for test-time denoising. Unlike prior methods that mainly target offline pre-training for high-level 3D tasks, our method enforces consistency across sample-specific noisy-noisier augmentations at test time, enabling the model to learn noise-invariant representations without architectural modifications.

\section{Method} 
\label{sec:method}

\subsection{Problem Formulation}

In the training phase, we consider a point cloud denoising model, $f_{\theta}: X \rightarrow Y$, designed to map a noisy point cloud $X$ to its clean counterpart $Y$. 
The model's parameters $\theta$ are optimized over a source-domain dataset $\mathcal{D}_s = \{(X^i_s, Y^i)\}_{i=1}^N$ by minimizing a supervised loss:
\begin{equation}
\min_{\theta} \mathbb{E}_{(X_s,Y)\sim\mathcal{D}_s}\left[\mathcal{L}_{\text{sup}}(f_{\theta}(X_s), Y)\right],
\end{equation}
where $X_s = Y + \epsilon_s$ is the input corrupted by synthetic noise $\epsilon_s \sim p_s$, and $\mathcal{L}_{\text{sup}}(\cdot)$ is a task-specific loss, typically measuring geometric discrepancy (e.g., point-wise Euclidean distance).

In the test phase, the pre-trained model $f_\theta$ encounters a new noisy input $X_t = Y + \epsilon_t$, where the noise $\epsilon_t$ is drawn from an unseen distribution $p_t$ that differs from the source distribution ($p_t \ne p_s$). 
This distribution shift introduces a significant domain gap, leading to a clear drop in performance when the model faces noise characteristics for which it was not trained.

Our goal is to adapt the pre-trained model $f_\theta$ to a new noisy point cloud $X_t$ with unseen noise, without access to any ground-truth $Y$ during the test phase. 
To accomplish this, we formulate this adaptation as an optimization problem, guided by a self-supervised loss on the test input:
\begin{equation}
\label{eq:optimization}
    \theta^* = \arg\min_{\theta} \mathbb{E} \left[ \mathcal{L}_{\text{self}}(f_\theta(X_t)) \right],
\end{equation}
where $\mathcal{L}_{\text{self}}(\cdot)$ is a self-supervised objective that leverages only the information contained within the test data itself to guide the parameter updates.

\subsection{Overview}

\begin{figure*}[!t]
    \centering
    \includegraphics[width=\textwidth]{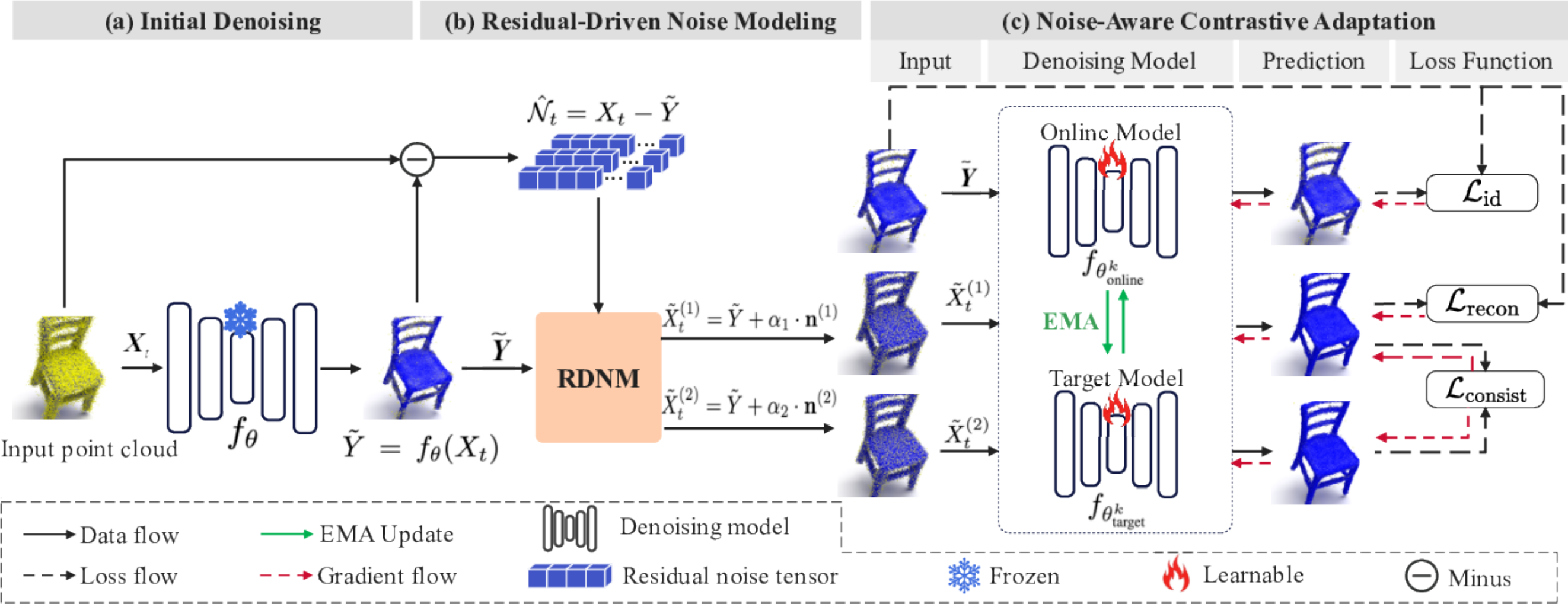}
    \caption{Overview of the proposed TTAD framework.
            (a) The input noisy point cloud $X_t$ is first denoised by a pre-trained model $f_\theta$ to produce a pseudo-clean output $\tilde{Y}$.
            (b) The residual-driven noise modeling (RDNM) module utilizes the residual noise $\hat{\mathcal{N}}_t = X_t - \tilde{Y}$ to estimate the test-time noise distribution $p(\mathbf{n})$ of $X_t$ via a lightweight Gaussian mixture model (GMM).
            (c) The noise-aware contrastive adaptation (NACA) module generates a sample-specific noisy-noisier pair $(\tilde{X}_t^{(1)}, \tilde{X}_t^{(2)})$ of $\tilde{Y}$.
            Within each test-time adaptation step, we first use the reconstruction loss $\mathcal{L}_{\text{recon}}$, identity loss $\mathcal{L}_{\text{id}}$, and consistency loss $\mathcal{L}_{\text{consist}}$ to update the online model $f_{\theta_{\text{online}}}$ via gradient descent at each iteration step.
            After that, the target model's parameters $\theta_{\text{target}}$ are updated as an exponential moving average (EMA) of the online model's parameters $\theta_{\text{online}}$. 
            Finally, we can use the final $\theta_{\text{target}}$ to obtain the adapted denoised result $Y_c = f_{\theta_{\text{target}}}(X_t)$.}
    \label{fig:framework}
\end{figure*}

The proposed TTAD framework facilitates a pre-trained point cloud denoising model to adapt to unseen noise during inference, without requiring clean supervision or retraining from source-domain data, as illustrated in Fig.~\ref{fig:framework}.
It consists of two main components: residual-driven noise modeling (RDNM) and noise-aware contrastive adaptation (NACA).
Specifically, given a test noisy point cloud $X_t$ and a pre-trained denoising model $f_\theta$, firstly, the RDNM estimates a sample-specific noise distribution $p(\mathbf{n})$ by analyzing the deviation between $X_t$ and its initial denoised prediction $\tilde{Y}$.
Then, NACA creates sample-specific noisy-noisier test-time training pair from $\tilde{Y}$, and adapts $f_\theta$ towards improved performance within a few adaptation steps using contrastive learning at test time. 
To stabilize the optimization and prevent overfitting to individual test samples, TTAD employs an exponential moving average (EMA) strategy to update model parameters $f_\theta$. 
Furthermore, TTAD naturally extends to a continual TTAD (CTTAD) setting, where the model parameters are progressively updated across sequential noisy point clouds characterized by a consistent noise distribution.
Overall, the proposed TTAD is architecture-agnostic and can be seamlessly integrated into various existing denoising models, enhancing their generalization ability under noise distribution shift at deployment.

\subsection{Residual-driven Noise Modeling}
\label{subsec:noise_modeling}

The residual-driven noise modeling (RDNM) aims to capture the complex and diverse noise characteristics of the input noisy point cloud $X_t$ based on a Gaussian mixture model at test-time.
Unlike existing denoising methods that assume sample-agnostic and synthetic noise in TTA-based image denoising \cite{vaksman2020lidia,fahim2023ss} and self-supervised point cloud denoising \cite{wang2024noise4denoise,su2025sitf}, our approach estimates the latent noise distribution directly from the input data in a sample-specific manner, resulting in more accurate and adaptive perturbation construction for test-time adaptation.

Firstly, given a test noisy sample $X_t \in \mathbb{R}^{3 \times N}$, we employ the pre-trained denoising model $f_\theta(\cdot)$ to produce a pseudo-clean point cloud $\tilde{Y}$, and compute the residual noise $\hat{\mathcal{N}}_t$:
\begin{equation}
\hat{\mathcal{N}}_t = X_t - \tilde{Y} ,
\end{equation}
where $\tilde{Y} = f_\theta(X_t)$, and $\hat{\mathcal{N}}_t \in \mathbb{R}^{3 \times N}$ consists of $N$ three-dimensional noise vectors $\{\mathbf{n}_i\}_{i=1}^N$.

Then, we fit a $M$-component Gaussian mixture model (GMM) to approximate the noise distribution of $X_t$:
\begin{equation}
\label{eq:gmm}
p(\mathbf{n}) = \sum_{m=1}^M \pi_m \mathcal{N}(\mathbf{n}; \boldsymbol{\mu}_m, \boldsymbol{\Sigma}_m),
\end{equation}
where $\pi_m$ are the mixture weights satisfying $\sum \pi_m = 1$, $\boldsymbol{\mu}_m$ and $\boldsymbol{\Sigma}_m$ denote the mean and covariance of the $m$-th Gaussian component, respectively.

The RDNM strategy allows the generation of sample-specific noise perturbations $p(\mathbf{n})$ that align with the actual noise distribution of the test sample, thereby providing reliable support for contrastive adaptation in downstream learning.

\subsection{Noise-Aware Contrastive Adaptation} 
\label{subsec:contrastive_adaptation}

The noise-aware contrastive adaptation (NACA) enables $f_\theta$ to adapt in a self-supervised manner by creating sample-specific noisy--noisier test-time training pair and enforcing output consistency using contrastive learning. 
Unlike existing TTA-based image denoising approaches which rely on synthetic or predefined noise assumptions (e.g., Gaussian noise)~\cite{fahim2023ss,su2025sitf}, the proposed NACA can capture sample-specific noise distributions encountered in real-world scenarios, improving the generalization performance under unseen noise at test time.
To this end, firstly, a sample-specific noisy-noisier pairing mechanism produces noisy-noisier pair $({\tilde{X}_t^{(1)}, \tilde{X}_t^{(2)}})$ that reflect the test sample's noise characteristics. 
Then, an exponential moving average (EMA) update strategy is employed to stabilize the online adaptation process while preserving the pre-trained model's performance on the source domain. 

\subsubsection{Sample-specific Noisy–Noisier Pairing}

For effective contrastive adaptation, we adopt a sample-specific noisy–noisier pairing strategy to create noisy–noisier test-time training pair $({\tilde{X}_t^{(1)}, \tilde{X}_t^{(2)}})$ based on the noise distribution $p(\mathbf{n})$ of $X_t$ estimated by RDNM.

Given the noise distribution $p(\mathbf{n})$ and pseudo-clean point cloud $\tilde{Y}$, we sample two independent noise vectors $\mathbf{n}^{(1)} \sim p(\mathbf{n})$, $\mathbf{n}^{(2)} \sim p(\mathbf{n})$ from $p(\mathbf{n}) $, and construct a noisy–noisier pair $({\tilde{X}_t^{(1)}, \tilde{X}_t^{(2)}})$ as:
\begin{equation}
\begin{split}
\tilde{X}_t^{(1)} = \tilde{Y} + \alpha_1 \cdot \mathbf{n}^{(1)}, \\
\tilde{X}_t^{(2)} = \tilde{Y} + \alpha_2 \cdot \mathbf{n}^{(2)},
\end{split}
\end{equation}
where $0 < \alpha_1 < \alpha_2 \leq 1$.
This pairing strategy enforces output consistency across varying noise levels, compelling the denoising model to learn noise-invariant feature representations. 
Moreover, the deliberately structured diversity in the input space enhances the model's robustness to distribution shifts, leading to superior generalization capability.

\subsubsection{Parameter Update}
\label{subsubsec:Parameter Update}

Updating model parameters $\theta$ during test-time adaptation must balance two competing objectives: adapting to unseen noise while preserving the denoising capability learned from the source domain.
This becomes particularly challenging when the noisy-noisier pair introduces large variations in the input, potentially leading to unstable optimization and catastrophic forgetting.
To mitigate these issues, we employ an exponential moving average (EMA) strategy to regularize the adaptation process and stabilize the learning dynamics throughout test-time inference.

Specifically, we introduce the online model $f_{\theta_{\text{online}}}$ and the target model $f_{\theta_{\text{target}}}$, both initialized from the pre-trained denoising model $f_{\theta}$.
$f_{\theta_{\text{online}}}$ and $f_{\theta_{\text{target}}}$ share identical architectures but differ in their parameters.
Given the initial denoised point cloud $\tilde{Y}$ and its noisy–noisier pair $({\tilde{X}_t^{(1)}, \tilde{X}_t^{(2)}})$, for the $k$-th adaptation step, we first set $\theta_{\text{online}}^{k} = \theta_{\text{target}}^{k-1}$.
Then, we employ the online model $f_{\theta_{\text{online}}^{k}}$ to update its parameters $\theta_{\text{online}}^{k}$ via gradient descent at each iteration, according to the loss function $\mathcal{L}_{\text{total}}$ discussed in Section \ref{subsubsec:loss}.
After that, the target model's parameters $\theta_{\text{target}}^{k}$ are updated as an exponential moving average of the online model's parameters $\theta_{\text{online}}^{k}$: 
\begin{equation}\label{eq:ema}
\theta_{\text{target}}^{k} = \beta \cdot \theta_{\text{target}}^{k-1} + (1 - \beta) \cdot \theta_{\text{online}}^{k},
\end{equation}
where $\beta \in [0,1)$ is the decay factor controlling the update smoothness.
The final denoised output $Y_c$ can be obtained by:
\begin{equation}
Y_c = f_{\theta_{\text{target}}^{k}}(X_t).
\end{equation}

This strategy offers two primary benefits.
First, it suppresses abrupt parameter fluctuations caused by noisy gradients, improving stability in the absence of supervision.
Second, by maintaining a trajectory-smoothed version of the model, EMA helps retain prior knowledge from the source domain and acts as a safeguard against overfitting to individual test samples.

\subsubsection{Loss Function}
\label{subsubsec:loss}

A key challenge in test-time adaptive denoising lies in designing an optimization objective that facilitates effective adaptation without access to clean supervision.
When an unseen noisy point cloud $X_t$ arrives at test time, we update the parameters of the online denoising model $f_{\theta_{\text{online}}}$ by minimizing a self-supervised loss, as ground-truth labels are unavailable in this setting.
Concretely, our objective combines three complementary terms: reconstruction loss, noise consistency loss, and identity constraint. 

\textbf{Reconstruction Loss} $\mathcal{L}_{\text{recon}}$.
This component encourages the denoising model to remain close to its pre-trained behavior by aligning its output $f_{\theta_{\text{online}}}(\tilde{X}_t^{(1)})$ on the noisy input $\tilde{X}_t^{(1)}$ with the pseudo-clean prediction $\tilde{Y}$, which serves as a soft supervision signal.
To maintain compatibility with different denoising methods, we directly use their predefined reconstruction metrics as the form of $\mathcal{L}_{\text{recon}}$.

\textbf{Noise Consistency Loss} $\mathcal{L}_{\text{consist}}$.
To promote robustness against varying noise intensities, we enforce consistency between the denoised outputs of the noisy–noisier input pair $(\tilde{X}_t^{(1)}, \tilde{X}_t^{(2)})$.
We use the symmetric Chamfer Distance (CD)~\cite{fan2017point} to measure geometric alignment:
\begin{equation}
\mathcal{L}_{\text{consist}}(A,B) = \frac{1}{|A|}\sum_{a\in A}\min_{b\in B}\|a-b\|^2 + \frac{1}{|B|}\sum_{b\in B}\min_{a\in A}\|b-a\|^2,
\end{equation}
where $A = f_{\theta_{\text{online}}}(\tilde{X}_t^{(1)})$, $B = f_{\theta_{\text{online}}}(\tilde{X}_t^{(2)})$.
This regularization encourages the network to learn noise-invariant representations and generalize across a broader range of test-time perturbations.

\textbf{Identity Constraint} $\mathcal{L}_{\text{id}}$.
To mitigate catastrophic forgetting during adaptation, we introduce a regularization term that preserves the model's behavior on its own predictions.
Specifically, the adapted model $f_{\theta_{\text{online}}}$ is encouraged to act as an identity mapping when applied to the pseudo-clean input:
\begin{equation}
\mathcal{L}_{\text{id}} = \|f_{\theta_{\text{online}}}(\tilde{Y}) - \tilde{Y}\|_2^2.
\end{equation}
This constraint maintains structural integrity and prevents the model from overfitting to perturbations introduced during adaptation.

The final loss function $\mathcal{L}_{\text{total}}$ is defined as:
\begin{equation}
\mathcal{L}_{\text{total}} = \mathcal{L}_{\text{recon}} + \lambda_1 \mathcal{L}_{\text{consist}} + \lambda_2 \mathcal{L}_{\text{id}},
\label{eq:loss}
\end{equation}
where $\lambda_1$ and $\lambda_2$ are hyperparameters controlling the relative importance of the consistency and identity terms, respectively.
The complete adaptation procedure of TTAD is summarized in Algorithm~\ref{alg:tta_denoise}.

\begin{algorithm}[!t]
\caption{Test-Time Adaptive Denoising for Point Cloud}
\label{alg:tta_denoise}
\begin{algorithmic}[1]
\REQUIRE Pre-trained denoising model $f_\theta$, input noisy point cloud $X_t$, learning rate $\eta$, EMA decay factor $\beta$, adaptation step $K$
\ENSURE Adapted denoised output $Y_c$

    \STATE Initialize $\theta_{\text{target}}^0 \gets \theta$
    \STATE Pseudo-clean point cloud: $\tilde{Y} = f_\theta(X_t)$
    \STATE Estimate residual noise: $\hat{\mathcal{N}}_t = X_t - \tilde{Y}$
    \STATE Fit Gaussian Mixture Model $p(\mathbf{n})$ to $\hat{\mathcal{N}}_t$
    \STATE Sample $n^{(1)} \sim p(\mathbf{n})$, $n^{(2)} \sim p(\mathbf{n})$
    \STATE Generate sample-specific noisy–noisier pair: \\
    \hspace{1.5em} $\tilde{X}^{(1)}_t = \tilde{Y} + \alpha_1 n^{(1)}$, \quad $\tilde{X}^{(2)}_t = \tilde{Y} + \alpha_2 n^{(2)}$
    \FOR{$k = 1$ to $K$}
        \STATE Initialize online weights $\theta_{\text{online}}^k \gets \theta_{\text{target}}^{k-1}$
        \STATE Compute total loss: $\mathcal{L}_{\text{total}} = \mathcal{L}_{\text{recon}} + \lambda_1 \mathcal{L}_{\text{consist}} + \lambda_2 \mathcal{L}_{\text{id}}$
        \STATE Update $\theta_{\text{online}}^k$ via gradient descent: \\
        \hspace{1.5em} $\theta_{\text{online}}^k \gets \theta_{\text{online}}^k - \eta \cdot \nabla_\theta \mathcal{L}_{\text{total}}$
        \STATE Update $\theta_{\text{target}}^k$ with EMA: \\
        \hspace{1.5em} $\theta_{\text{target}}^k \gets \beta \theta_{\text{target}}^{k-1} + (1-\beta) \theta_{\text{online}}^k$
    \ENDFOR
    \STATE Final output: $Y_c = f_{\theta_{\text{target}}^k}(X_t)$
\end{algorithmic}
\end{algorithm}

\subsection{Continual TTAD}

The CTTAD is proposed to improve adaptation efficiency for scenarios where point clouds encounter noise distributions with consistent characteristic.
In the standard TTAD setup, the adaptation for each test sample \( X_t \) is performed independently by solving the optimization problem defined in Equation~(\ref{eq:optimization}), with model parameters always initialized from the pre-trained source model \( \theta \).
After adaptation, the updated parameters \( \theta^* \) are discarded, and the next test sample is adapted from the same initialization.
For sequential test samples with shared noise distributions, CTTAD achieves continual adaptation by progressively updating model parameters, eliminating the need for per-sample resets.

Specifically, for each test sample $X_t^i$ from a target-domain dataset $\mathcal{D}_t$, the denoising model is initialized with the parameters adapted from the previous sample $X_t^{i-1}$:
\begin{equation}
\theta_{i} \gets \theta_{i-1}.
\end{equation}
This permits the model to accumulate knowledge progressively from past samples \( \{X_t^1, X_t^2, \dots, X_t^{i-1}\} \), thereby improving its generalization to future inputs.
To balance accuracy and efficiency, the number of adaptation steps $K_i$ is determined by an annealing schedule:
\begin{equation}
K_i = \texttt{anneal}(i, K) = \max(0, \lfloor \gamma^{i-1} \cdot K \rfloor),
\end{equation}
where \(\gamma \in (0,1)\) controls the decay rate. 
The denoising model then performs TTAD on \(X_t^i\) for \(K_i\) adaptation steps using the self-supervised objective in Equation~(\ref{eq:loss}). 
Finally, the denoised point cloud is obtained by \(Y_{c,i} = f_{\theta_{\text{target},i}}(X_t^i)\). 
The CTTAD gradually reduces the number of adaptation steps over time, mirroring a test-time annealing process, which helps stabilize the model as it processes more test samples.
The procedure of CTTAD is elaborated in Algorithm~\ref{alg:ctta}.

\begin{algorithm}[!t]
\caption{Continual Test-Time Adaptive Denoising for Point Cloud}
\label{alg:ctta}
\begin{algorithmic}[1]
\REQUIRE Initial model parameters $\theta$; adaptation step $K$; sequence of test point clouds $X_t^i$ from the target-domain dataset $\mathcal{D}_t$
\ENSURE Adapted predictions $\{Y_{c, i}\}$ and parameter trajectory $\{\theta_i\}$

\STATE Initialize: $\theta_{\text{online}, 0} \gets \theta$, $\theta_{\text{target}, 0} \gets \theta$
\FOR{each input noisy point cloud $X_t^i \in \mathcal{D}_t$}
    \STATE Inherit parameters from the previous sample $X_t^{i-1}$: \\
    \hspace{1.5em} $\theta_{\text{online}, i} \gets \theta_{\text{online}, i-1}$, \quad $\theta_{\text{target}, i} \gets \theta_{\text{target}, i-1}$
    \STATE Determine adaptation step: $K_i = \texttt{anneal}(i,\ K)$
    \STATE Perform TTAD on $X_t^i$ for $K_i$ steps 
    \STATE Output denoised result $Y_{c, i} = f_{\theta_{\text{target}, i}}(X_t^i)$ and $\theta_i = \theta_{\text{target}, i}$
\ENDFOR
\end{algorithmic}
\end{algorithm}

\section{Experiment} 
\subsection{Experimental Setup}
\label{subsec:setup}

\subsubsection{Datasets}

We evaluate the proposed TTAD framework on both synthetic and real-world datasets to comprehensively assess its performance under Gaussian and diverse unseen noise conditions.

For experiments on synthetic datasets, we follow standard protocols and adopt the widely used PUNet dataset~\cite{yu2018pu}, originally designed for point cloud upsampling and denoising.
To evaluate generalization under distribution shift, five types of unseen noise are introduced at test time:
(1) \textit{Pulse noise}, which randomly replaces a subset of points with large displacements;
(2) \textit{Exponential noise}, which introduces long-tailed, asymmetric perturbations;
(3) \textit{Laplace noise}, which produces sharp and heavy-tailed perturbations;
(4) \textit{Uniform noise}, which samples perturbations from a bounded uniform distribution; and
(5) \textit{Mixed noise}, which combines Gaussian, Laplace, and Uniform perturbations to simulate more complex hybrid corruptions.
These noise types differ significantly from the Gaussian distribution assumed during training and are designed to challenge the generalization ability of denoising models under diverse distribution shifts.
For this setting, we sample 20 unseen CAD meshes from the PUNet dataset~\cite{yu2018pu} and generate noisy point clouds at 50K resolution.
Each model is evaluated both with and without the TTAD framework to quantify its effectiveness in improving generalization under domain shift.
This setup ensures sufficient noise diversity while providing a more comprehensive evaluation of robustness to unseen noise distributions.

For real-world evaluation, we employ the Paris-Rue-Madame dataset~\cite{serna2014paris}, which consists of LiDAR scans of outdoor urban street scenes in Paris.
This dataset features various real-world sensor noise patterns, including motion artifacts, occlusions, and multipath interference, making it a suitable benchmark for evaluating generalization under naturally occurring corruptions.
As the ground-truth geometry is not available, we conduct qualitative comparisons by visually inspecting the denoised outputs.

\subsubsection{Denoising Models}

To verify the generality of our approach, we apply TTAD to three representative point cloud denoising backbones:
\begin{itemize}
    \item ScoreDenoise (Score)~\cite{luo2021score}: A score-based denoising model that applies score matching to approximate the gradient of the data distribution for sampling-based restoration.
    \item PointFilter (PF)~\cite{zhang2020pointfilter}: A spatial filtering-based method that predicts residual corrections for noisy inputs by aggregating local neighborhood information.
    \item IterativePFN (IPFN)~\cite{de2023iterativepfn}: An iterative refinement network that leverages feature fusion and confidence weighting to progressively restore clean point clouds.
\end{itemize}

To provide a stronger and more comprehensive evaluation, we further include two recent and closely related denoising methods:
\begin{itemize}
    \item LaPDA~\cite{LaPDA2025}: A recent point cloud denoising method that enhances denoising performance through input adaptation at test time together with adaptive geometry-aware feature modeling.
    \item PD-LTS~\cite{Mao_2024_CVPR}: A latent-space point cloud denoising method that leverages graph convolution and invertible neural networks to disentangle noise from geometric structure.

\end{itemize}

We use official pre-trained weights for ScoreDenoise and IterativePFN. 
For PointFilter, we retrain the denoising model using its public implementation and recommended hyperparameters on the PUNet training set~\cite{yu2018pu}. 
For LaPDA and PD-LTS, we use their public implementations and report the corresponding comparison results under our experimental setting.
For a fair comparison, all iterative baseline methods are configured to use 2 inference iterations unless otherwise specified.
Specifically, 20 CAD meshes are selected, and Poisson disk sampling is applied to generate uniformly distributed point clouds at three resolutions: 10K, 30K, and 50K.
To construct noisy inputs for training, we add zero-mean Gaussian noise with standard deviations randomly sampled from the range of 0.5\% to 2.0\% of the object's bounding sphere radius.

\subsubsection{Adaptation Details}

For all TTAD and CTTAD experiments, we use the Adam optimizer with learning rate of $1 \times 10^{-4}$.
The number of component $M$ of GMM is set as $5$.
The loss weights are set as $\lambda_1 = 0.2$ for the consistency term and $\lambda_2 = 0.1$ for the identity regularization.
The EMA decay factor is fixed at $\beta = 0.99$.
In the standard TTAD setting, each test sample is adapted for $K = 100$ iterations.
For noisy-noisier perturbations, we use $\alpha_1 = 0.3$ and $\alpha_2 = 0.8$ to generate contrastive input pair with distinct noise intensities.
In the CTTAD setting, $\gamma$ is set as 0.7.
All experiments are performed on an NVIDIA RTX 4090 GPU.

\subsubsection{Evaluation Protocol}  

We evaluate the denoising performance under both synthetic and real-world noise settings. 
Chamfer Distance (CD)~\cite{fan2017point} and Point-to-Mesh distance (P2M)~\cite{li2021point} are adopted as quantitative metrics.
For real world validation, we conduct qualitative analysis on the Paris-Rue-Madame (PRM) dataset~\cite{serna2014paris}. 
As ground truth is unavailable, visual comparisons are used to assess the perceptual quality of denoised outputs.

\subsection{Comparison on Synthetic Dataset}

\subsubsection{Quantitative Results}

\begin{table}[!t]
    \centering
    \caption{Performance comparison of denoising methods on the PUNet dataset under in-distribution Gaussian noise. 
    ``Train/Test'' denotes the noise type used for training and testing, respectively. \textbf{Bold} and \underline{underlined} values indicate the best and second-best performance, respectively. CD and P2M values are multiplied by $10^4$.}
    \resizebox{0.8\linewidth}{!}{
    \begin{tabular}{c|cc|cc}
        \toprule
        \multirow{2}{*}{Method} 
        & \multicolumn{2}{c|}{1\% Gaussian/Gaussian} 
        & \multicolumn{2}{c}{2\% Gaussian/Gaussian} \\
        & CD$\downarrow$ & P2M$\downarrow$ & CD$\downarrow$ & P2M$\downarrow$ \\
        \midrule
        Score      & 0.86 & 0.55 & 2.10 & 1.68 \\
        + TTAD     & 0.90 & 0.58 & 1.66 & 1.26 \\
        + CTTAD    & 0.89 & 0.57 & 1.74 & 1.28 \\
        \midrule
        PF         & 0.80 & 0.42 & 1.49 & 1.07 \\
        + TTAD     & 0.82 & 0.44 & 1.27 & 0.91 \\
        + CTTAD    & 0.81 & 0.43 & 1.19 & 0.84 \\
        \midrule
        IPFN       & 0.61 & 0.32 & 1.31 & 0.88 \\
        + TTAD     & 0.62 & \underline{0.31} & 1.15 & 0.81 \\
        + CTTAD    & 0.62 & \textbf{0.29} & \underline{1.08} & \textbf{0.75} \\
        \midrule
        LaPDA      & \underline{0.57} & 0.35 & 1.71 & 1.26 \\
        PD-LTS     & \textbf{0.54} & 0.32 & \textbf{0.97} & \underline{0.76} \\
        \bottomrule
    \end{tabular}
    }
    \label{tab:gaussian_only}
\end{table}

\begin{table*}[!t]
    \centering
    \caption{Performance comparison of denoising methods on the PUNet dataset under unseen synthetic noise. ``Train/Test'' denotes the noise type used for training and testing, respectively. \textbf{Bold} and \underline{underlined} values indicate the best and second-best performance, respectively. CD and P2M values are multiplied by $10^4$.}
    \resizebox{\textwidth}{!}{
    \begin{tabular}{c|cccccccc|cccccccc}
        \toprule
        \multirow{2}{*}{Method}
        & \multicolumn{8}{c|}{1\%}
        & \multicolumn{8}{c}{2\%} \\
        & \multicolumn{2}{c}{Gaussian/Pulse}
        & \multicolumn{2}{c}{Gaussian/Exponential}
        & \multicolumn{2}{c}{Gaussian/Laplace}
        & \multicolumn{2}{c|}{Gaussian/Uniform}
        & \multicolumn{2}{c}{Gaussian/Pulse}
        & \multicolumn{2}{c}{Gaussian/Exponential}
        & \multicolumn{2}{c}{Gaussian/Laplace}
        & \multicolumn{2}{c}{Gaussian/Uniform} \\
        & CD$\downarrow$ & P2M$\downarrow$
        & CD$\downarrow$ & P2M$\downarrow$
        & CD$\downarrow$ & P2M$\downarrow$
        & CD$\downarrow$ & P2M$\downarrow$
        & CD$\downarrow$ & P2M$\downarrow$
        & CD$\downarrow$ & P2M$\downarrow$
        & CD$\downarrow$ & P2M$\downarrow$
        & CD$\downarrow$ & P2M$\downarrow$ \\
        \midrule

        Score      & 1.49 & 1.13 & 2.80 & 2.03 & 0.96 & 0.60 & 1.98 & 1.33 & 4.60 & 4.40 & 7.71 & 6.67 & 2.23 & 1.79 & 10.45 & 9.31 \\
        + TTAD     & 1.10 & 0.74 & 2.55 & 1.90 & 0.95 & 0.62 & 1.32 & 0.86 & 3.72 & 3.50 & 6.20 & 5.62 & 1.46 & 1.10 & 8.15 & 6.66 \\
        + CTTAD    & 1.12 & 0.77 & 2.37 & 1.77 & 0.95 & 0.62 & 1.33 & 0.89 & 3.66 & 3.52 & 5.64 & 5.17 & 1.53 & 1.12 & 7.41 & 6.13 \\
        \midrule

        PF         & 1.12 & 0.80 & 2.40 & 1.84 & 1.06 & 0.65 & 1.49 & 0.93 & 3.42 & 3.26 & 6.15 & 4.99 & 2.44 & 2.06 & 5.03 & 4.33 \\
        + TTAD     & 0.96 & 0.64 & 2.26 & 1.79 & 1.10 & 0.69 & 1.21 & 0.74 & \underline{1.60} & 2.22 & 4.10 & 3.34 & 2.37 & 1.97 & \underline{3.38} & \underline{2.73} \\
        + CTTAD    & \underline{0.90} & \underline{0.59} & \underline{2.12} & \textbf{1.68} & 1.03 & 0.64 & 1.13 & 0.68 & \textbf{1.49} & 2.04 & 3.77 & 3.07 & 2.20 & 1.81 & \textbf{3.12} & \textbf{2.51} \\
        \midrule

        IPFN       & 1.40 & 0.88 & 2.66 & 1.95 & \underline{0.70} & \textbf{0.43} & \underline{0.96} & \underline{0.61} & 3.98 & 3.71 & 5.47 & 4.24 & 1.22 & 0.88 & 15.84 & 13.52 \\
        + TTAD     & 1.12 & 0.75 & 2.41 & 1.90 & 0.74 & \underline{0.46} & 0.97 & \underline{0.61} & 2.01 & \underline{1.77} & \underline{3.13} & \underline{2.86} & \underline{1.18} & \underline{0.85} & 15.56 & 13.24 \\
        + CTTAD    & 1.09 & 0.70 & 2.27 & 1.75 & 0.74 & \textbf{0.43} & \textbf{0.95} & \textbf{0.57} & 1.95 & \textbf{1.63} & \textbf{2.98} & \textbf{2.68} & \textbf{1.11} & \textbf{0.79} & 14.81 & 12.41 \\
        \midrule

        LaPDA      & 0.94 & 0.67 & \textbf{2.07} & \underline{1.69} & \underline{0.70} & 0.47 & 1.81 & 1.18 & 5.25 & 5.24 & 6.38 & 5.44 & 2.59 & 2.17 & 11.38 & 10.01 \\
        PD-LTS     & \textbf{0.80} & \textbf{0.54} & 2.40 & 1.95 & \textbf{0.66} & \textbf{0.43} & 1.42 & 0.93 & 3.68 & 3.57 & 7.10 & 5.96 & 1.33 & 1.05 & 13.07 & 10.99 \\
        \bottomrule
    \end{tabular}
    }
    \label{tab:unseen_main}
\end{table*}

Table~\ref{tab:gaussian_only} reports the in-distribution results under Gaussian noise, where the test corruption is consistent with the source-domain noise used during training. 
In this setting, the three baseline denoising backbones already achieve relatively strong performance, and the improvements brought by TTAD and CTTAD are generally smaller than those under unseen noise distributions. 
This observation is expected, since test-time adaptation is primarily designed to handle distribution shifts rather than further optimize already matched source-target conditions. 
Nevertheless, TTAD and CTTAD usually maintain performance comparable to their corresponding backbones under Gaussian noise, while still yielding moderate gains at the higher noise level. 
For example, under 2\% Gaussian noise, PF-CTTAD reduces CD from 1.49 to 1.19 and P2M from 1.07 to 0.84, while IPFN-CTTAD reduces CD from 1.31 to 1.08 and P2M from 0.88 to 0.75. 

Compared with LaPDA and PD-LTS, our TTAD-enhanced models remain competitive under Gaussian noise, although the relative advantage is less pronounced than that under unseen settings. 
Overall, these results indicate that the proposed adaptation framework does not significantly compromise the original denoising ability when no distribution shift is present.

Table~\ref{tab:unseen_main} reports the quantitative results under four single unseen noise distributions, including Pulse, Exponential, Laplace, and Uniform noise. 
When the test noise deviates from the Gaussian corruption used during training, all three baseline denoising backbones suffer from clear performance degradation, confirming the severity of domain shift in point cloud denoising. 
Among these unseen settings, Pulse and Exponential noise generally lead to the most pronounced deterioration, while Laplace and Uniform noise reveal different sensitivities of the competing methods to heavy-tailed and bounded perturbations, respectively.

Incorporating TTAD or CTTAD consistently improves the robustness of the three backbones under most single unseen noise conditions, with the gains being particularly evident at the higher noise level of 2\%. 
For example, under 2\% Pulse noise, PF-CTTAD reduces CD from 3.42 to 1.49 and P2M from 3.26 to 2.04, while under 2\% Exponential noise, IPFN-CTTAD reduces CD from 5.47 to 2.98 and P2M from 4.24 to 2.68. 
 
Compared with LaPDA and PD-LTS, our method remains competitive and often achieves better results when combined with suitable backbones. 
In particular, PD-LTS attains the best performance under several relatively mild 1\% settings, such as Pulse and Laplace noise, indicating the advantage of structured feature modeling when the corruption level is moderate. 
However, as the noise level increases, backbones with TTAD enhancement, especially PF-CTTAD and IPFN-CTTAD, exhibit stronger robustness and achieve the best overall performance in most 2\% single unseen-noise settings. 
These results suggest that the proposed test-time adaptation strategy becomes increasingly effective as the test corruption departs further from the source-domain Gaussian distribution.

Table~\ref{tab:mixed_only} further reports the results under the mixed noise setting, where Gaussian, Laplace, and Uniform perturbations are combined to simulate more complex hybrid corruptions. 
Compared with a single synthetic noise type, this setting is more challenging because the corruption no longer follows a homogeneous distribution and instead contains multiple perturbation patterns simultaneously. 
Under 1\% mixed noise, PD-LTS achieves the best CD, while IPFN and PD-LTS obtain the best P2M, suggesting that structured representation learning is effective when the hybrid corruption remains relatively moderate. 
However, under the more challenging 2\% mixed noise, IPFN-CTTAD achieves the best performance, with CD and P2M reduced to 1.73 and 1.24, respectively, outperforming both the original IPFN and the additional comparison baselines. 
This result demonstrates that the proposed continual adaptation strategy can better handle complex hybrid corruptions by progressively adjusting the denoiser to the sample-specific noise characteristics observed at test time. 
Overall, the mixed-noise results provide further evidence that our method is not limited to a single unseen distribution, but can also generalize effectively to more realistic settings where different perturbation types coexist.

\begin{table}[!t]
    \centering
    \caption{Performance comparison of denoising methods on the PUNet dataset under mixed unseen noise. ``Train/Test'' denotes the noise type used for training and testing, respectively. \textbf{Bold} and \underline{underlined} values indicate the best and second-best performance, respectively. CD and P2M values are multiplied by $10^4$.}
    \resizebox{0.8\linewidth}{!}{
    \begin{tabular}{c|cc|cc}
        \toprule
        \multirow{2}{*}{Method}
        & \multicolumn{2}{c|}{1\% Gaussian/Mixed}
        & \multicolumn{2}{c}{2\% Gaussian/Mixed} \\
        & CD$\downarrow$ & P2M$\downarrow$ & CD$\downarrow$ & P2M$\downarrow$ \\
        \midrule
        Score      & 1.04 & 0.64 & 3.36 & 2.80 \\
        + TTAD     & 0.99 & 0.63 & 1.84 & 1.40 \\
        + CTTAD    & 0.97 & 0.63 & \underline{1.80} & 1.36 \\
        \midrule
        PF         & 1.01 & 0.59 & 2.26 & 1.84 \\
        + TTAD     & 1.03 & 0.62 & 2.03 & \underline{1.34} \\
        + CTTAD    & 0.98 & 0.58 & 1.88 & 1.49 \\
        \midrule
        IPFN       & \underline{0.70} & \textbf{0.42} & 1.98 & 1.48 \\
        + TTAD     & 0.75 & 0.46 & 1.82 & \underline{1.34} \\
        + CTTAD    & 0.73 & \underline{0.43} & \textbf{1.73} & \textbf{1.24} \\
        \midrule
        LaPDA      & 0.72 & 0.46 & 3.44 & 2.86 \\
        PD-LTS     & \textbf{0.66} & \textbf{0.42} & 1.89 & 1.48 \\
        \bottomrule
    \end{tabular}
    }
    \label{tab:mixed_only}
\end{table}

\subsubsection{Qualitative Results}  

\begin{figure*}[!t]
    \centering
    \begin{subfigure}{0.98\textwidth}
        \centering
        \includegraphics[width=\textwidth]{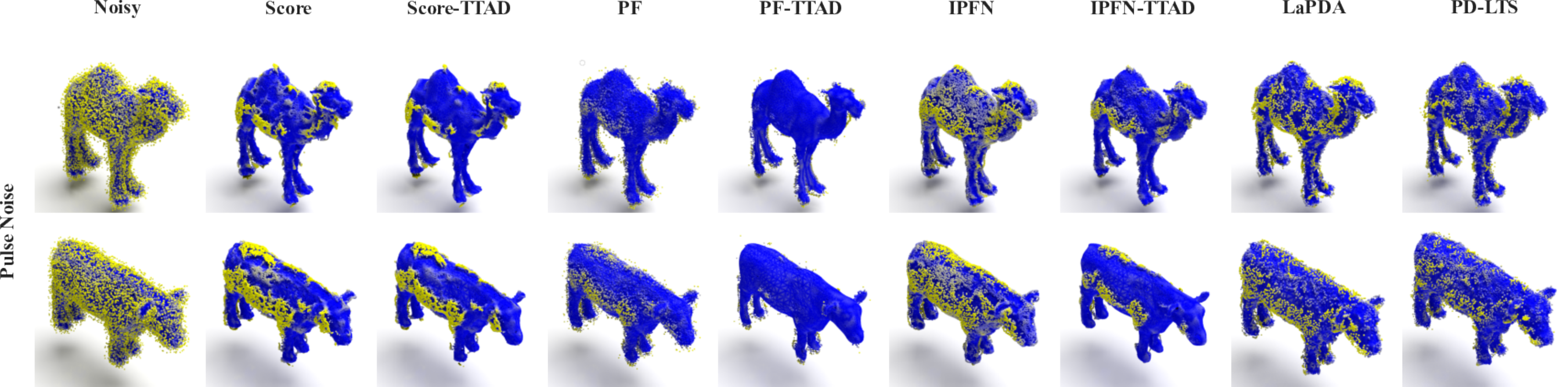}
        \caption{Qualitative comparison under Pulse noise.}
        \label{fig:vis_pulse}
    \end{subfigure}
    
    \vspace{1mm}
    
    \begin{subfigure}{0.98\textwidth}
        \centering
        \includegraphics[width=\textwidth]{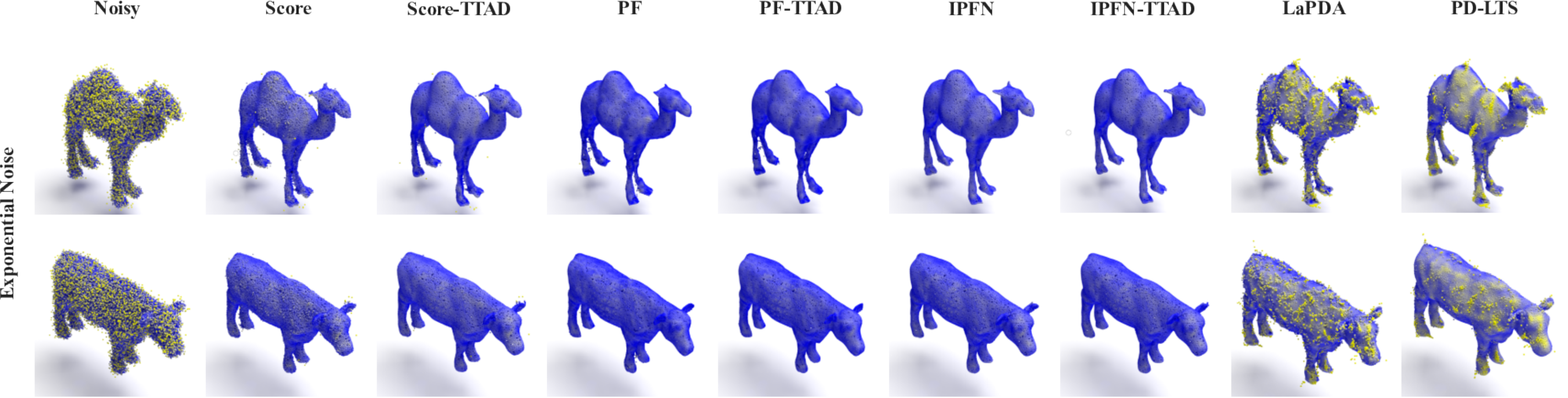}
        \caption{Qualitative comparison under Exponential noise.}
        \label{fig:vis_exponential}
    \end{subfigure}

    \caption{
    Visualization of denoising results on different unseen 2\% noise conditions.
    All baseline models (Score~\cite{luo2021score}, PF~\cite{zhang2020pointfilter}, IPFN~\cite{de2023iterativepfn}, LaPDA~\cite{LaPDA2025}, and PD-LTS~\cite{Mao_2024_CVPR}) are trained on Gaussian noise and tested on inputs corrupted by (a) Pulse noise and (b) Exponential noise. 
    }
    \label{fig:qualitative_all_1}
\end{figure*}

\begin{figure*}[!t]
    \centering
    \begin{subfigure}{0.98\textwidth}
        \centering
        \includegraphics[width=\textwidth]{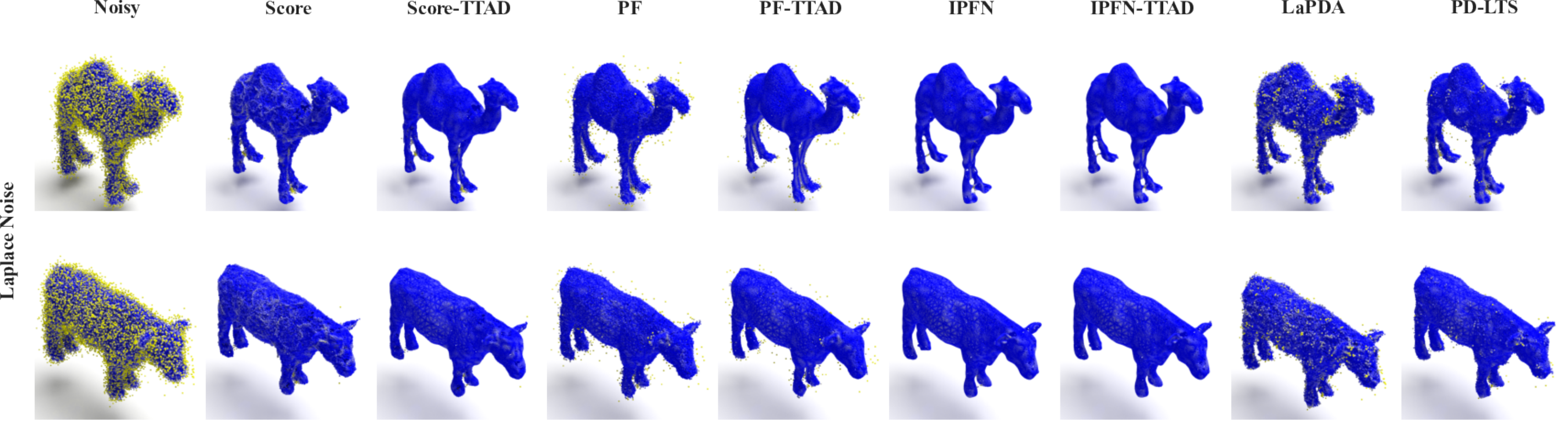}
        \caption{Qualitative comparison under Laplace noise.}
        \label{fig:vis_laplace}
    \end{subfigure}
    
    \vspace{1mm}
    
    \begin{subfigure}{0.98\textwidth}
        \centering
        \includegraphics[width=\textwidth]{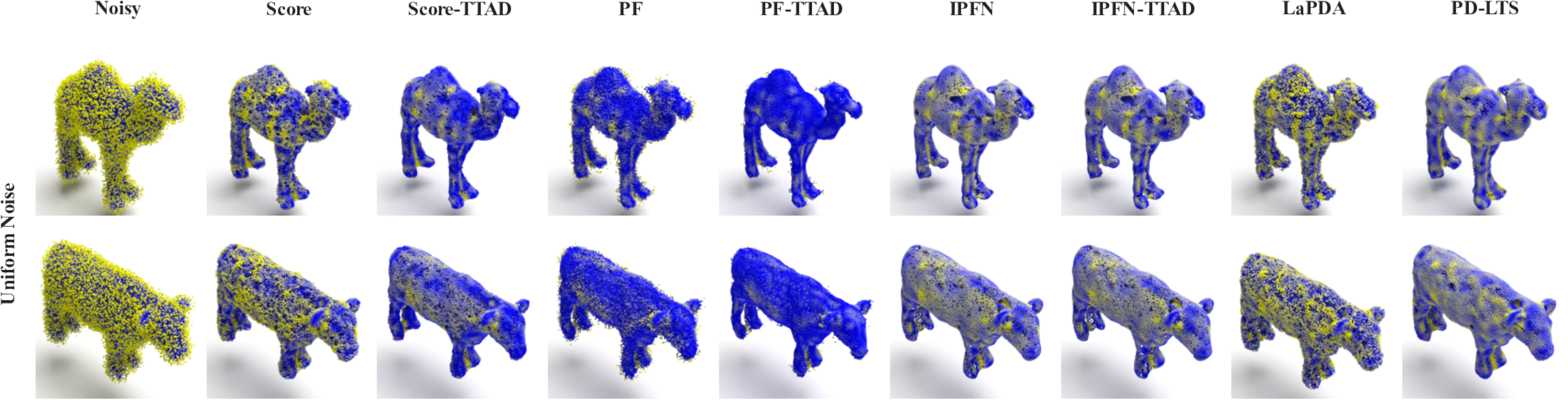}
        \caption{Qualitative comparison under Uniform noise.}
        \label{fig:vis_uniform}
    \end{subfigure}
    
    \vspace{1mm}
    
    \begin{subfigure}{0.98\textwidth}
        \centering
        \includegraphics[width=\textwidth]{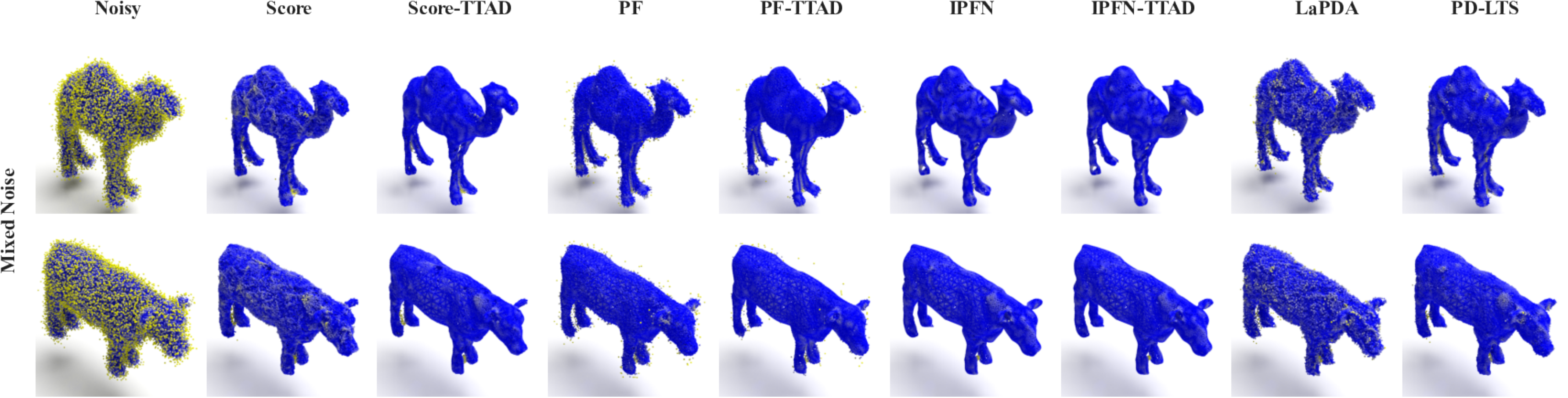}
        \caption{Qualitative comparison under Mixed noise.}
        \label{fig:vis_mixed}
    \end{subfigure}
    
    \caption{
    Visualization of denoising results on additional unseen 2\% noise conditions. 
    All baseline models (Score~\cite{luo2021score}, PF~\cite{zhang2020pointfilter}, IPFN~\cite{de2023iterativepfn}, LaPDA~\cite{LaPDA2025}, and PD-LTS~\cite{Mao_2024_CVPR}) are trained on Gaussian noise and tested on inputs corrupted by (a) Laplace noise, (b) Uniform noise, and (c) Mixed noise. 
    }
    \label{fig:qualitative_all_2}
\end{figure*}

Figure~\ref{fig:qualitative_all_1} and Figure~\ref{fig:qualitative_all_2} present visual comparisons between baseline models and their TTAD-enhanced counterparts under five unseen 2\% noise conditions, including Pulse, Exponential, Laplace, Uniform, and Mixed noise.
Without adaptation, the original denoising backbones often exhibit noticeable artifacts, such as residual outliers, clustered noise, oversmoothed local structures, and incomplete geometric details, especially as the noise level increases.
After incorporating TTAD, these artifacts are visibly alleviated, resulting in cleaner point distributions, smoother surfaces, and more complete structural boundaries.

More specifically, under Pulse noise, PF with TTAD effectively removes sparse outliers and yields a cleaner overall shape, while under Exponential noise, IPFN with TTAD shows improved preservation of geometric continuity and surface smoothness.
Under Laplace and Uniform noise, TTAD-enhanced models generally retain sharper contours and more stable local geometry than their corresponding original backbones.
For the more challenging Mixed noise, the benefit of test-time adaptation becomes even more evident, as the TTAD-enhanced results exhibit stronger robustness to hybrid perturbations and better preserve the overall object structure.

Compared with the additional baselines LaPDA and PD-LTS, our TTAD-enhanced models achieve competitive visual quality across all five noise types.
LaPDA shows favorable preservation of geometric features, while PD-LTS, as a flow-based latent-space method, generally produces smooth and coherent denoised surfaces.
This comparison reflects the trade-off between representation capacity and adaptation flexibility: latent-space methods provide structured representations for geometry--noise disentanglement, whereas TTAD estimates sample-specific residual perturbations to adapt existing Euclidean-space denoisers at test time.
As shown in the visual results, this plug-and-play adaptation strategy effectively improves the original backbones under diverse unseen noise conditions while maintaining competitive visual quality against recent latent-space baselines.
These observations qualitatively support the quantitative findings.

\subsection{Comparison on Real-world Dataset}
\begin{figure*}[!t]
    \centering
    \includegraphics[width=\linewidth]{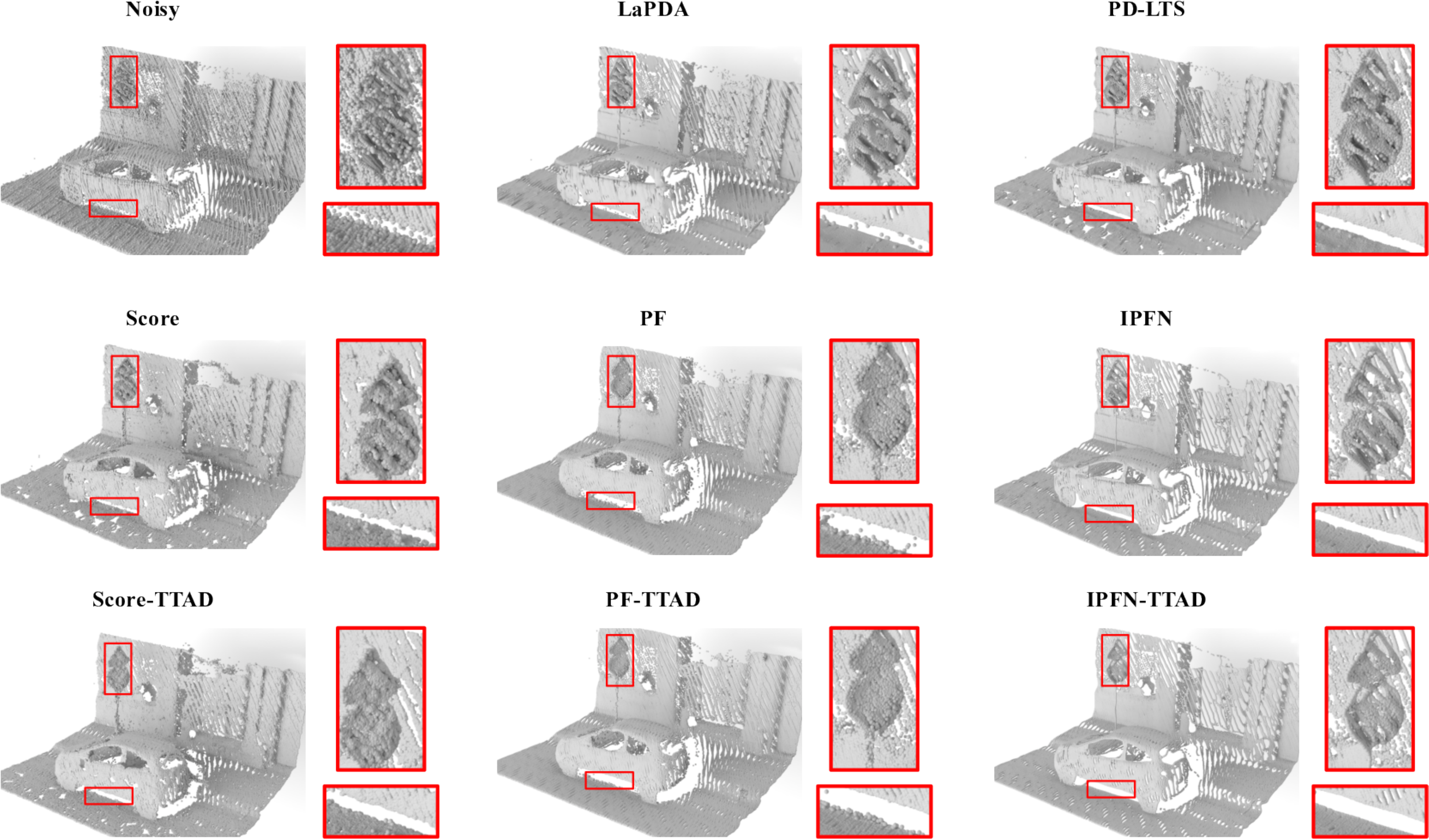}
    \caption{ Qualitative results on the Paris-Rue-Madame dataset \cite{serna2014paris}. TTAD improves denoising quality under real-world noise, removing scattered outliers, suppressing LiDAR scan-line artifacts, and restoring smooth and coherent surface structures.}
    \label{fig:TTAD_Madame}
\end{figure*}

We further evaluate the TTAD framework on the Paris-Rue-Madame (PRM) dataset~\cite{serna2014paris}, which contains LiDAR scans of urban street scenes with diverse real-world noise patterns, including motion-induced distortions, ghosting artifacts, background clutter, and non-uniform sampling. 
Since clean ground-truth point clouds are unavailable, we provide qualitative comparisons in Figure~\ref{fig:TTAD_Madame}, where two representative regions are further enlarged for detailed inspection.

As shown in Figure~\ref{fig:TTAD_Madame}, the selected scene includes a roadside sign affected by surrounding clutter and ghosting artifacts. 
Non-adaptive baseline models such as PF~\cite{zhang2020pointfilter} and ScoreDenoise~\cite{luo2021score} exhibit noticeable surface roughness and residual scattered noise, especially in the highlighted regions. 
IPFN~\cite{de2023iterativepfn} produces relatively smooth outputs, but local structural details can still be further refined.

In contrast, incorporating TTAD consistently improves the visual quality across the three backbones. 
The TTAD-enhanced results show cleaner point distributions, reduced noise accumulation, and more clearly defined structural boundaries. 
In particular, PF-TTAD effectively suppresses scattered outliers, while IPFN-TTAD better preserves the continuity of geometric structures, as evidenced by the smoother contours and more coherent surfaces in the zoomed-in regions.

The comparison methods, LaPDA~\cite{LaPDA2025} and PD-LTS~\cite{Mao_2024_CVPR}, also produce visually competitive denoising results. 
LaPDA shows favorable preservation of geometric features, while PD-LTS generates smooth and coherent surfaces through latent-space modeling. 
Compared with these recent baselines, the TTAD-enhanced models achieve comparable visual quality while maintaining strong adaptability to the sample-specific real-world noise patterns in the test scene.

These improvements can be attributed to the two key components of our framework. 
The RDNM module simulates realistic test-time perturbations to expose the model to sample-specific noise patterns, while the NACA strategy enforces consistency across these perturbations. 
Together, they guide the model to learn noise-invariant geometric features, enabling effective adaptation to complex real-world noise without requiring clean supervision.

\subsection{Computational Efficiency}

\begin{figure}[!t]
    \centering
    \includegraphics[width=\linewidth]{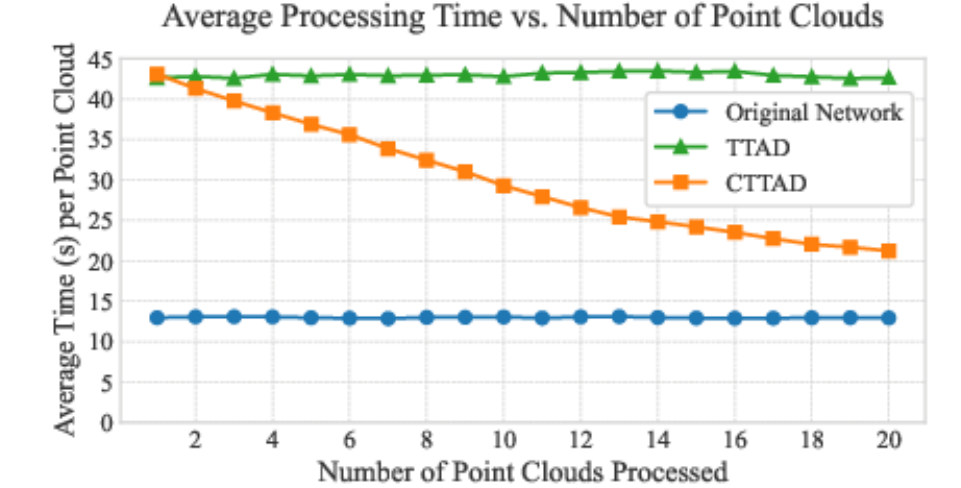}
    \caption{
    Average processing time comparison. 
    The baseline model (blue) has a constant low runtime. 
    Standard TTAD (green) incurs a constant high cost due to per-sample re-initialization. 
    In contrast, CTTAD (orange) becomes increasingly efficient by reusing and incrementally updating parameters across samples.}
    \label{fig:ctta_time}
\end{figure}

We first evaluate the computational efficiency by measuring the average processing time per point cloud under the pulse noise setting, where the reported values are averaged over three TTAD-compatible denoising backbones. 
As shown in Figure~\ref{fig:ctta_time}, the analysis includes three methods: the original network without adaptation (Baseline),  TTAD (TTAD), and continual TTAD (CTTAD).

The baseline model maintains a low and constant processing time, as it only performs inference.
In contrast, the standard TTAD incurs a high and constant computational cost, as it performs a full, independent adaptation from scratch for each test sample.
The CTTAD, while initially incurring a similar cost to standard TTAD, progressively reduces its average processing time.
This significant efficiency gain is due to the parameter reuse mechanism: instead of re-initializing for each sample, the model reuses and incrementally updates the parameters learned from previous instances. 
This makes CTTAD particularly well-suited for online or streaming scenarios where test samples arrive continuously.

\begin{figure*}[!t] 
    \centering
    \begin{subfigure}[b]{0.49\linewidth}
        \includegraphics[trim=0 0 0 0, clip, width=\linewidth]{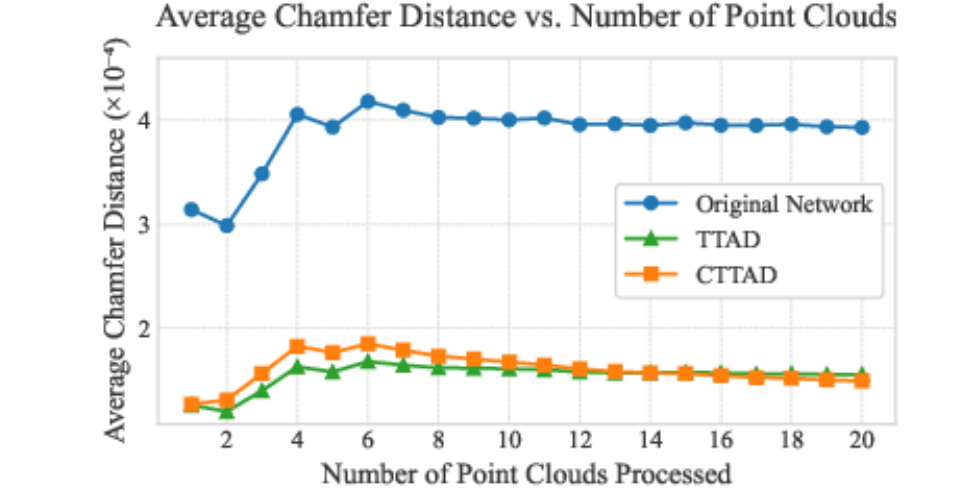} 
        \caption{Average Chamfer distance}
        \label{fig:cd_performance}
    \end{subfigure}
    \begin{subfigure}[b]{0.49\linewidth}
        \includegraphics[trim=0 0 0 0, clip, width=\linewidth]{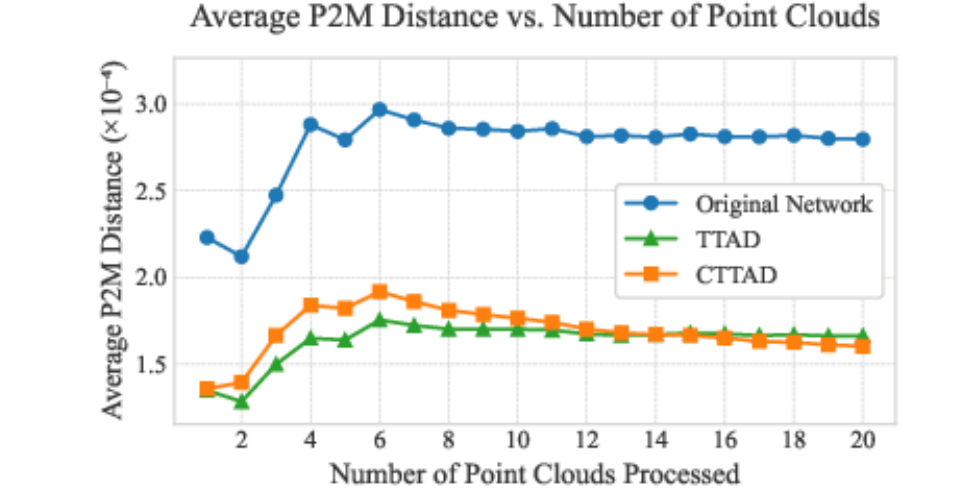} 
        \caption{Average P2M distance}
        \label{fig:p2m_performance}
    \end{subfigure}
    \caption{
    Performance comparison under unseen 2\% Pulse noise, showing the average error tracked over a sequence of point clouds. While TTAD (green) shows stable, strong performance, CTTAD (orange) initially trades a minor performance dip for long-term gain, ultimately outperforming both baselines.
    }
    \label{fig:progressive_performance}
\end{figure*}

Beyond efficiency, we then analyze the evolution of the average denoising error (Chamfer Distance and P2M Distance) as the adaptation process unfolds over a sequence of test samples under the unseen 2\% Pulse noise condition.
The performance dynamics, illustrated in Figure~\ref{fig:progressive_performance}, reveal the core advantage of our continual learning approach by tracking the average error of the three methods over a sequence of test samples.
The original network exhibits consistently high error, serving as a performance lower bound.
The TTAD achieves a strong, stable denoising performance.
It acts as a powerful baseline for single-instance adaptation but shows no improvement over time as it treats each sample independently. 
During the initial phase (processing the first few samples), CTTAD incurs a minor performance cost, with its error being slightly higher than that of the standard TTAD. 
This reflects a transient adaptation phase where the model builds its long-term memory.
However, as more samples are processed, the accumulated knowledge allows CTTAD to rapidly improve, eventually surpassing the standard TTAD and achieving the best overall performance.

In conclusion, these results validate our central hypothesis: by integrating knowledge from previous samples, CTTAD not only achieves superior amortized efficiency but also reaches a higher final denoising accuracy, proving its effectiveness for real-world, continuous data streams.

\subsection{Ablation Study}  
\label{subsec:Ablation}

\begin{table}[!t]
    \footnotesize
    \centering
    \caption{
    Ablation study on components in TTAD.
    Results are averaged over three selected baseline denoising methods.
    "NNP" denotes the noisy-noisier pairing strategy. 
    A checkmark (\checkmark) indicates the module is enabled, and a cross (\ding{55}) indicates it is disabled.
    CD and P2M values are multiplied by $10^4$.
    The best results under each noise type are highlighted in \textbf{bold}.
    }
    \begin{tabular}{c c c c c c c c}
        \toprule
        \multirow{2}{*}{RDNM} & \multirow{2}{*}{NNP} & \multicolumn{2}{c}{2\%Gaussian} & \multicolumn{2}{c}{2\%Pulse} & \multicolumn{2}{c}{2\%Exponential} \\ \cmidrule{3-8}
         & & CD & P2M & CD & P2M & CD & P2M \\ \midrule
        \ding{55} & \ding{55} & 1.63 & 1.21 & 4.00 & 2.84 & 6.44 & 5.31 \\
        \ding{55} & \ding{51} & 1.42 & 1.25 & 4.12 & 4.57 & 5.56 & 6.00 \\
        \ding{51} & \ding{55} & 1.72 & 1.33 & 3.01 & 2.55 & 5.16 & 4.97 \\
        \ding{51} & \ding{51} & \textbf{1.36} & \textbf{0.99} & \textbf{2.44} & \textbf{2.50} & \textbf{4.47} & \textbf{3.94} \\ 
        \bottomrule
    \end{tabular}
    \label{tab:Ablation-GMM-DICP}
\end{table}

\subsubsection{Residual-driven Noise Modeling}  
We first assess the effectiveness of the RDNM module by comparing it against a simplified alternative that employs Gaussian noise to create noisy-noisier training pair during adaptation.

As shown in Table~\ref{tab:Ablation-GMM-DICP}, our RDNM consistently outperforms the Gaussian baseline across all synthetic noise types.
The performance gap is particularly pronounced under unseen noise conditions (e.g., exponential noise), where the simplistic Gaussian assumption fails to capture the specific characteristics of test-time corruption.
From the noise modeling perspective, Gaussian noise introduces spatially uniform and unimodal perturbations, limiting the diversity and realism of augmented inputs.
\begin{figure}[!t]
    \centering
    \includegraphics[width=\linewidth]{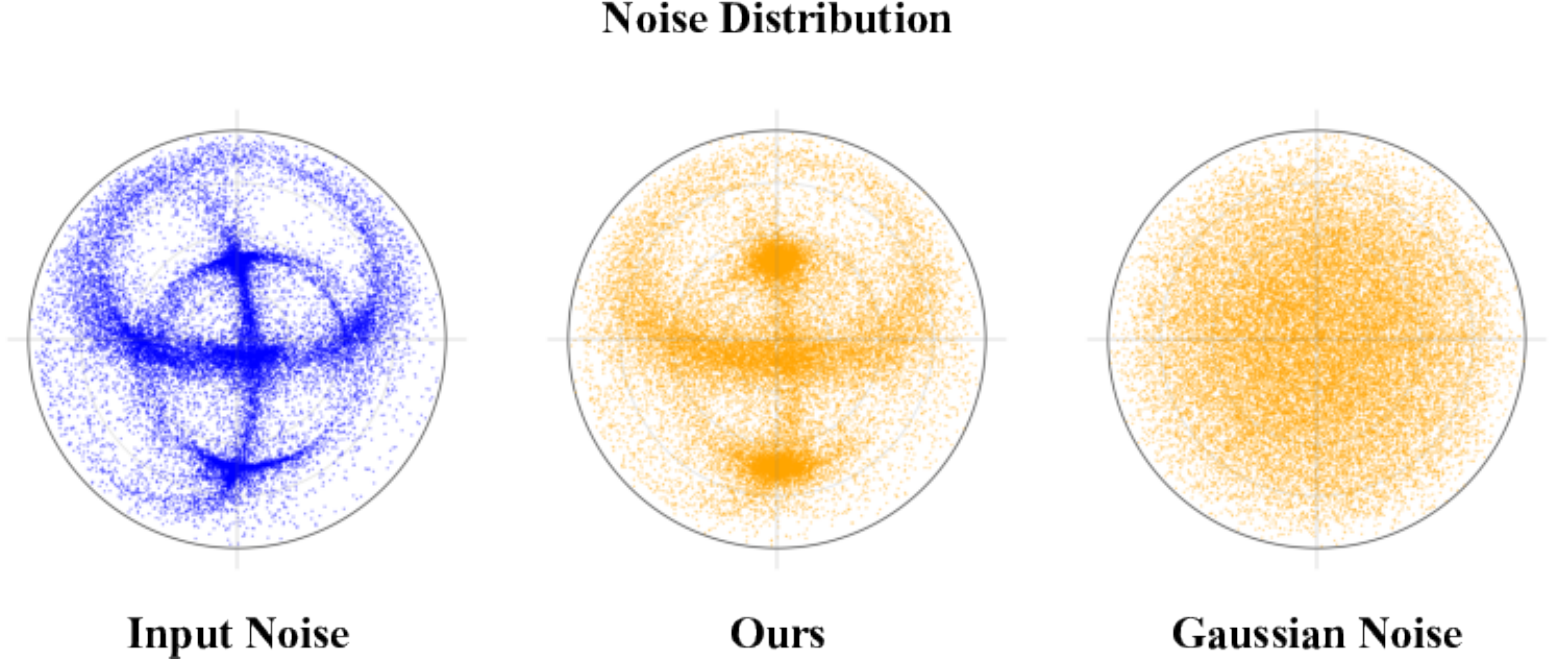}
    \caption{
    Visualization of noise direction distributions using spherical-to-polar projection.
    Left: Ground-truth input noise shows a structured pattern. 
    Middle: Our RDNM captures the dominant directional structures effectively. 
    Right: Standard Gaussian noise appears spatially uniform and lacks structure.
    Each dot represents a normalized noise direction projected to 2D; the denser areas indicate dominant noise directions.
    }
    \label{fig:noise_vis}
\end{figure}
In contrast, our approach estimates the actual noise distribution from residuals, which captures the multi-modal patterns critical for generating realistic test-time perturbations.

To better illustrate the differences in noise modeling, we visualize the directional distributions of three types of noise: the ground-truth input noise, our residual-driven noise sampled from the learned GMM, and standard Gaussian noise.
We compute the residual vectors for each point, normalize them to obtain unit vectors, and map them to a 2D polar plane using a spherical projection. 
As shown in Figure~\ref{fig:noise_vis}, the input noise exhibits a highly structured distribution, reflecting complex and directional corruption patterns.
Our modeled noise captures these directional structures with high fidelity, showing concentrated patterns that align closely with the ground-truth distribution.
In contrast, the Gaussian noise is spatially uniform and lacks the directional specificity of real test-time noise.
This confirms that our RDNM provides a more realistic and informative estimation of the actual noise distribution, which is critical for effective and robust test-time adaptation.

\begin{figure*}[!t]
    \centering
    \includegraphics[width=\linewidth]{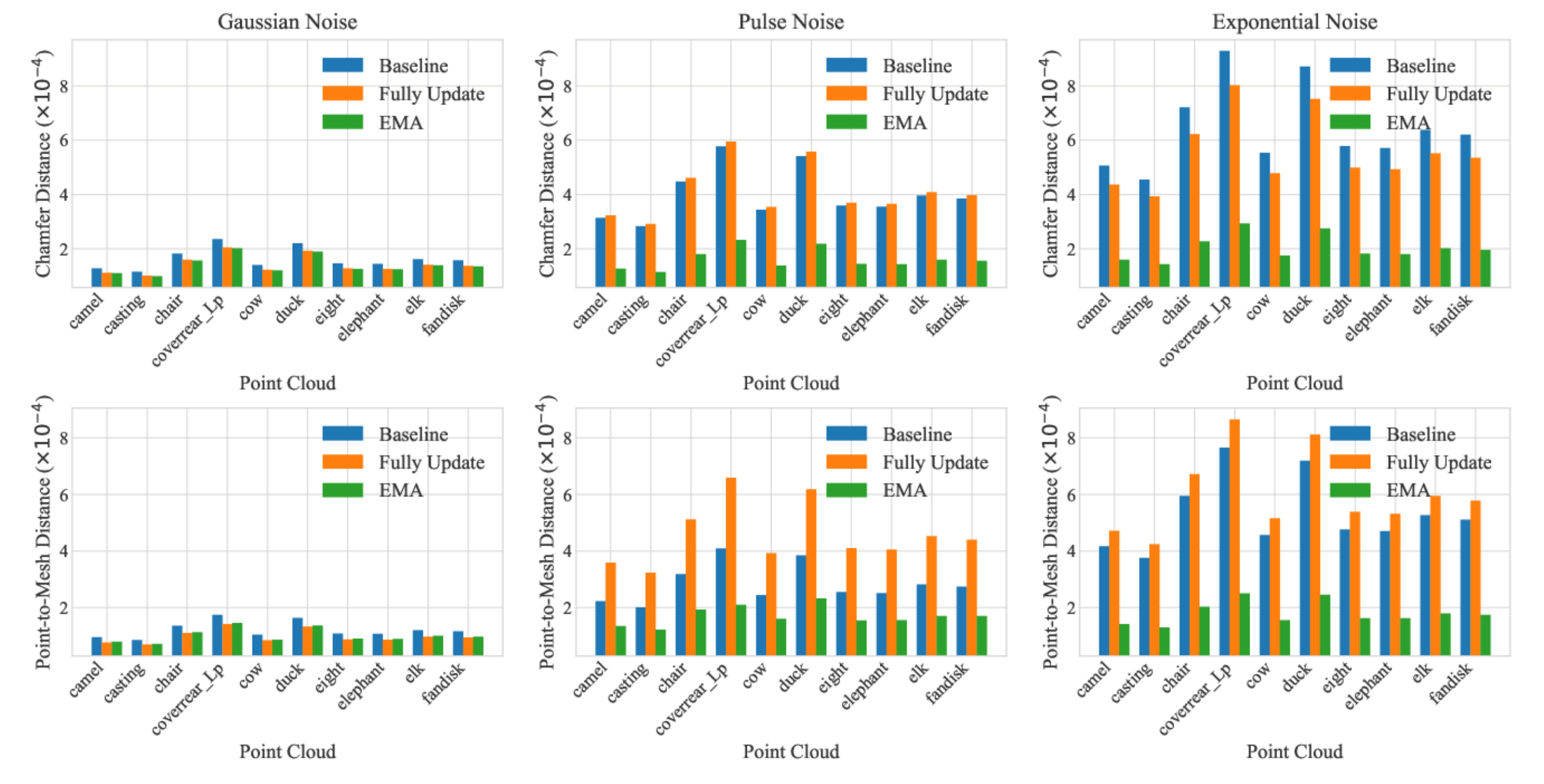}
    \caption{
    Performance comparison of different parameter update strategies on point clouds under Gaussian and unseen (Pulse, Exponential) synthetic noise. Results are averaged over three selected baseline denoising methods.
    Each group compares three strategies: Baseline (blue), Fully Update (orange), and EMA (green), evaluated by Chamfer Distance (top row) and Point-to-Mesh Distance (bottom row).
    Across all noise types, the adaptation strategy with EMA consistently improves generalization to unseen noise while maintaining source-domain performance on Gaussian noise.
}
    \label{fig:Ablation_EMA}
\end{figure*}

\subsubsection{Noisy–Noisier Pairing}  

To evaluate the impact of sample-specific noisy–noisier pair with varying perturbation intensities, we design a baseline method which perturbs each input sample only once with a fixed-strength noise for comparison.
Specifically, the consistency loss $\mathcal{L}_{\text{consist}}$, which relies on a pair of augmented inputs, is consequently deactivated.

As reported in Table~\ref{tab:Ablation-GMM-DICP}, our noisy-noisier pairing strategy consistently yields lower errors and improves generalization across all noise types compared to this baseline.
The mechanism's success can be attributed to the structured contrastive learning it enables.
The less-corrupted noisy sample serves to anchor the optimization around a reliable pseudo-clean reference.
Simultaneously, the more heavily-corrupted noisier counterpart introduces marked  contrastive variation.
By enforcing output consistency between these two versions via $\mathcal{L}_{\text{consist}}$, the model is compelled to learn features that are robust to varying noise intensities.
Essentially, our pairing mechanism enhances the adaptation process by providing structured data diversity, guiding the model toward learning truly noise-invariant representations.

\subsubsection{Parameter Update Strategy}  
We further assess the impact of our parameter update mechanism by comparing full gradient-based updates with the EMA strategy under 2\% noise setting, where the reported values represent the average performance over three TTAD-compatible denoising backbones.
As reported in Figure~\ref{fig:Ablation_EMA}, although both methods improve performance over the original pre-trained denoising models, EMA yields more robust results, particularly under unseen noise types such as pulse and exponential noise.
In addition, we test both variants under Gaussian noise, which is the same distribution used during training, to examine potential forgetting effects.
The EMA version retains significantly better performance, suggesting that it helps preserve the source-domain knowledge while adapting to new noise.
This supports our hypothesis that EMA acts as a regularizer, mitigating catastrophic forgetting by gradually incorporating new information without abruptly overwriting the learned priors.
Moreover, EMA behaves as a temporal ensemble over the online network trajectory, smoothing noisy gradients caused by imperfect pseudo clean references.
This stabilizes the adaptation process, especially when the pseudo-clean reference is inaccurate.

\subsubsection{Component Number of GMM}  
\begin{figure}[H]
    \centering
    \includegraphics[width=\columnwidth]{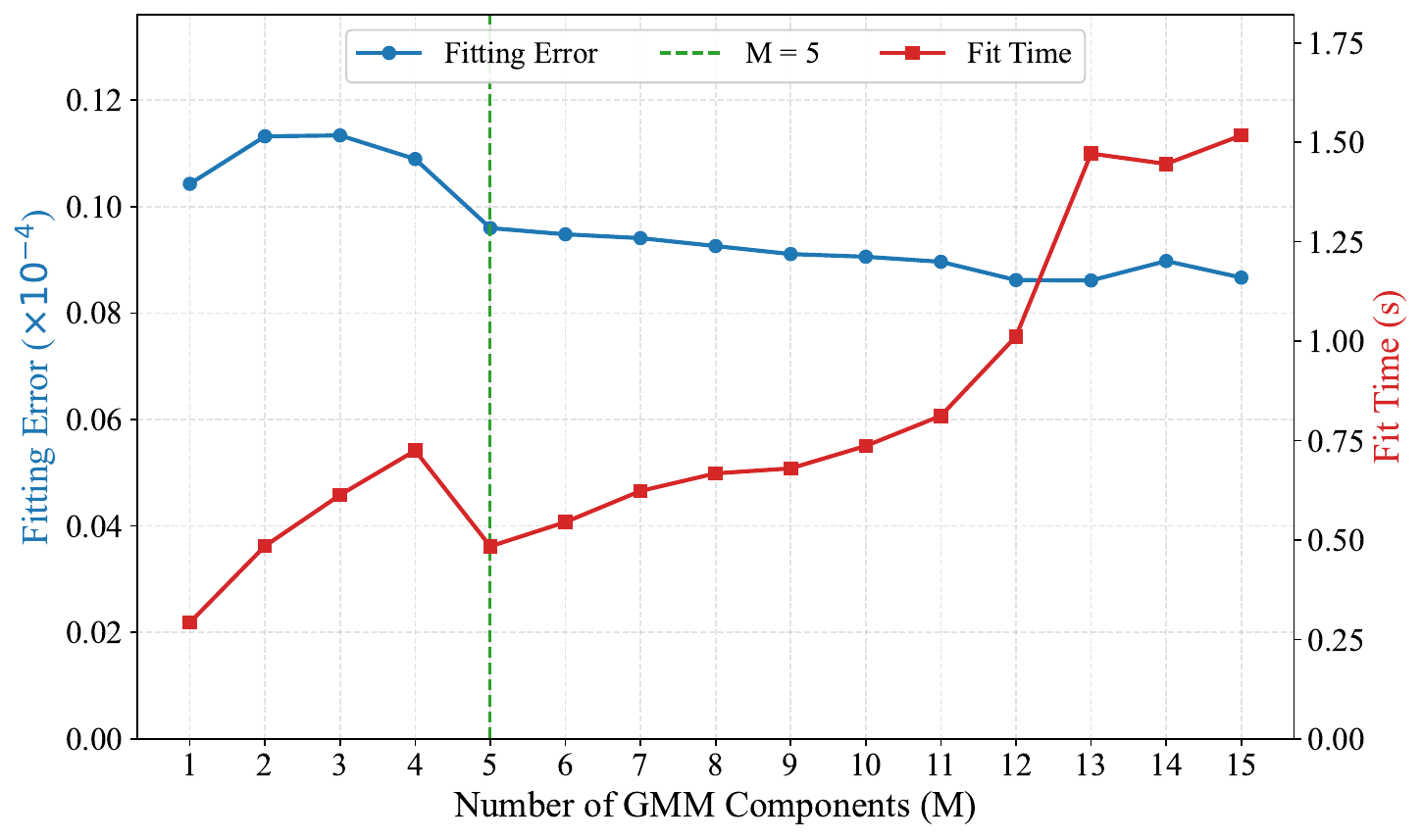} 
    \caption{Effect of GMM component number $M$ on the distribution fitting error (blue curve) and computational time (red curve) under complex mixed noise.}
    \label{fig:gmm_m}
\end{figure}

We further investigate the sensitivity of our residual-driven noise modeling to the number of Gaussian mixture components $M$. 
To this end, we evaluate the distribution Fitting Error (Chamfer Distance between the estimated GMM noise and the actual unseen mixed noise) and the corresponding Fit Time under varying $M \in[1, 15]$. 
As illustrated in Figure \ref{fig:gmm_m}, when $M < 5$, the GMM struggles to capture the complex, multi-modal nature of the mixed noise, yielding a large fitting error. 
Notably, at $M=5$, the fitting error experiences a significant drop and stabilizes. 
Beyond $M=5$, the fitting error plateaus with only marginal improvements, while the computational cost (Fit Time) grows substantially, particularly for $M > 10$.
Consequently, we set $M=5$ as the optimal trade-off between noise modeling accuracy and test-time computational efficiency.

\section{Conclusion and Future Work} 
\label{sec:conclusion}

We propose a Test-Time Adaptive Denoising (TTAD) framework to address the domain shift issue in point cloud denoising in this paper.
By leveraging residual-driven noise estimation, noise-aware contrastive adaptation, and EMA strategies, TTAD enables pre-trained denoising models to adapt to unseen noise distributions at test time, without requiring additional supervision or retraining from source data.
Extensive experiments demonstrate that TTAD consistently enhances the performance of state-of-the-art denoising models under diverse unseen noise conditions, achieving significant reductions in both Chamfer Distance and Point-to-Mesh metrics.
Additionally, we show that the EMA update strategy outperforms full parameter updates, preserving performance of denoising models across domains.
Beyond standard adaptation, our framework also supports a CTTAD mode that facilitates efficient online adaptation over data streams.
This extension further enhances its practicality for real-time point cloud processing tasks.

Future work will primarily focus on enhancing the long-term stability of CTTAD, a central challenge in Online Test-Time Adaptation (OTTA).
Our analysis revealed that when the underlying denoising model generates less reliable pseudo-clean references (an issue observed particularly with weaker baselines like ScoreDenoise \cite{luo2021score}), the continual adaptation process can be susceptible to error accumulation over long data streams.
To mitigate this, several avenues can be explored.
One promising direction is to introduce uncertainty-aware update mechanisms, where the model adapts more cautiously when its confidence in the estimated clean reference is low.
Another direction involves investigating strategic parameter reset policies or partial rollbacks to the original pre-trained state, preventing the model from drifting too far from its robust initial knowledge.

\section*{Acknowledgments}

The authors would like to acknowledge the helpful comments and kindly suggestions provided by anonymous referees.

\bibliographystyle{elsarticle-num} 
\bibliography{TTAD}

\end{document}